\pdfoutput=1
\documentclass[11pt]{article}
\usepackage{naacl2021}

\usepackage{times}
\usepackage{latexsym}
\usepackage{algorithmicx}
\usepackage{algpseudocode}
\usepackage{algorithm}
\usepackage{upgreek}

\usepackage{CJKutf8}
\usepackage{multirow}
\usepackage{microtype}
\usepackage{todonotes}
\usepackage{booktabs}
\usepackage{amsmath,amsfonts,bm}

\newcommand{\poly}{\textsc{PolyGloss}}
\newcommand{\bee}{\textsc{Bumblebee}}

\newcommand{\mytilde}{{\raise.17ex\hbox{$\scriptstyle\mathtt{\sim}$}}}

\def\eqref#1{equation~\ref{#1}}
\def\1{\bm{1}}

\DeclareMathAlphabet{\mathsfit}{\encodingdefault}{\sfdefault}{m}{sl}
\SetMathAlphabet{\mathsfit}{bold}{\encodingdefault}{\sfdefault}{bx}{n}

\def\gL{{\mathcal{L}}}
\def\gM{{\mathcal{M}}}

\def\gP{{\mathcal{P}}}

\def\sL{{\mathbb{L}}}

\def\sP{{\mathbb{P}}}

\def\sS{{\mathbb{S}}}

\DeclareMathOperator*{\argmax}{arg\,max}

\usepackage{amssymb}
\usepackage[nameinlink]{cleveref}
\usepackage{savesym}
\savesymbol{I}
\usepackage[utf8]{inputenc}
\usepackage[vietnamese,greek,russian,hindi,arabic,english]{babel}

\usepackage{devanagari}
\usepackage{subcaption}

\usepackage{anyfontsize}
\usepackage{enumitem}
\usepackage{soul}
\sethlcolor{pink}

\crefformat{section}{\S#2#1#3} % see manual of cleveref, section 8.2.1
\crefname{algorithm}{Alg.}{Algs.}
\crefformat{subsection}{\S#2#1#3}
\Crefname{equation}{Eq.}{Eqs.}
\Crefname{figure}{Fig.}{Figs.}

\definecolor{cornflowerblue}{HTML}{4a86e8}
\definecolor{myorange}{HTML}{ff9900}

\title{Code-Mixing on Sesame Street: Dawn of the Adversarial Polyglots}
\author{Samson Tan$^{\S\natural}$ \quad Shafiq Joty$^{\S\ddagger}$ \\
  $^\S$Salesforce Research Asia\quad
  $^\natural$National University of Singapore \quad
  $^\ddagger$Nanyang Technological University \\
  \texttt{\{samson.tan,sjoty\}@salesforce.com} \\
}

\date{}

\begin{document}
\maketitle
\begin{abstract}
%\vspace{-0.3em}
Multilingual models have demonstrated impressive cross-lingual transfer performance. However, test sets like XNLI are monolingual at the example level. In multilingual communities, it is common for polyglots to code-mix when conversing with each other. Inspired by this phenomenon, we present two strong black-box adversarial attacks (one word-level, one phrase-level) for multilingual models that push their ability to handle code-mixed sentences to the limit. The former uses bilingual dictionaries to propose perturbations and translations of the clean example for sense disambiguation. The latter directly aligns the clean example with its translations before extracting phrases as perturbations. Our phrase-level attack has a success rate of 89.75\% against XLM-R$_\text{large}$, bringing its average accuracy of 79.85 down to 8.18 on XNLI. Finally, we propose an efficient adversarial training scheme that trains in the same number of steps as the original model and show that it improves model accuracy.\footnote{Code: \href{https://github.com/salesforce/adversarial-polyglots}{github.com/salesforce/adversarial-polyglots}}
\end{abstract}

\section{Introduction}
%\vspace{-0.3em}
The past year has seen incredible breakthroughs in cross-lingual generalization with the advent of massive multilingual models that aim to learn universal language representations \citep{pires2019multilingualbert,wu2019beto,conneau-etal-2020-emerging}. These models have demonstrated impressive cross-lingual transfer abilities: simply fine-tuning them on task data from a high resource language such as English after pretraining on monolingual corpora was sufficient to manifest such abilities. This was observed even for languages with different scripts and no vocabulary overlap \citep{K2020Cross-Lingual}.

\begin{figure}[t]
    \begin{subfigure}{0.485\textwidth}
    \centering
    \includegraphics[width=\textwidth]{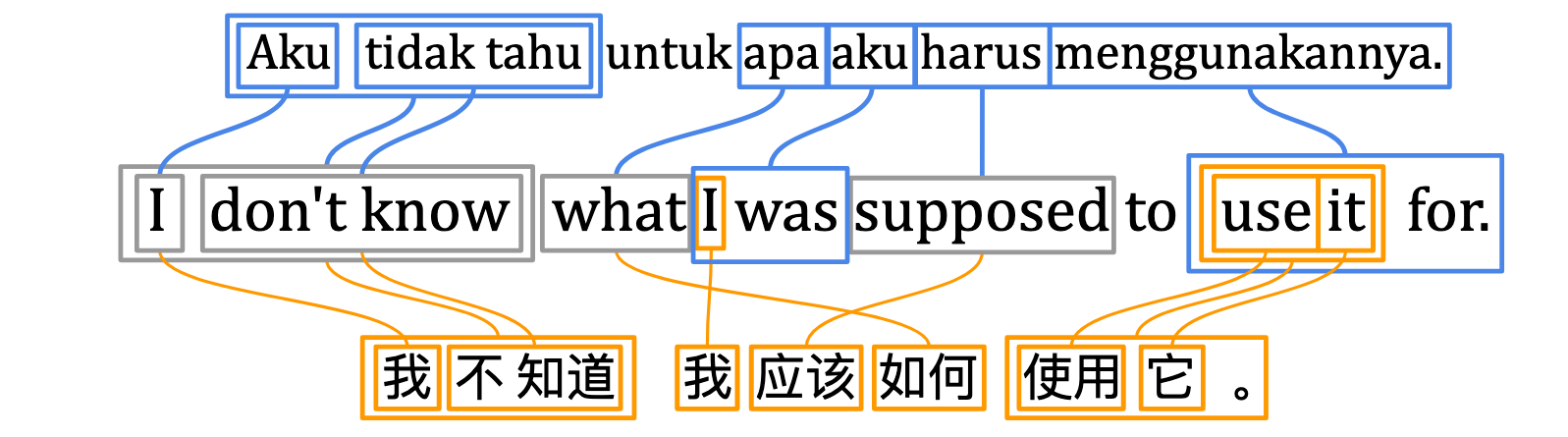}
    \vspace{-1.5em}
    \caption{Aligned words across sentences}
    \end{subfigure}
    \begin{subfigure}{0.485\textwidth}
    \centering
    %\vspace{1em}
    \includegraphics[width=\textwidth]{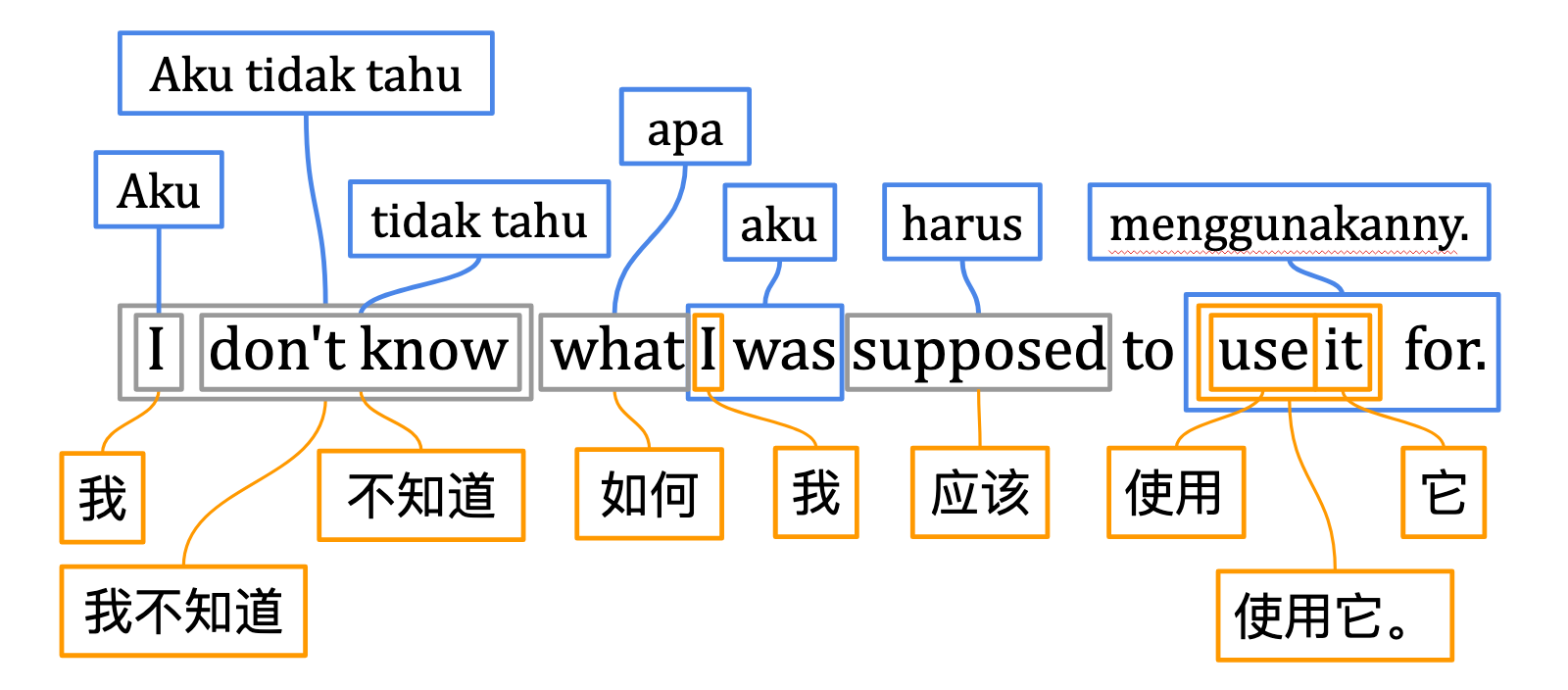}
    \vspace{-1.75em}
    \caption{Extracted candidate perturbations}
    \end{subfigure}
    \begin{subfigure}{0.485\textwidth}
    \centering
    \includegraphics[width=\textwidth]{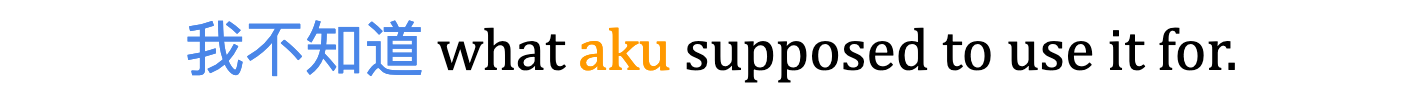}
    \vspace{-1.5em}
    \caption{Final multilingual adversary}
    \end{subfigure}
    \vspace{-1em}
    \caption{\bee's three key stages of adversary generation: (a) Align words in the matrix (English) and embedded sentences (top: \textcolor{cornflowerblue}{\textbf{Indonesian}}, bottom: \textcolor{myorange}{\textbf{Chinese}}); (b) Extract candidate perturbations from embedded sentences; (c) Construct final adversary by maximizing the target model's loss.}
    \vspace{-0.5em}
    \label{fig:bumblebee}
\end{figure}{}

However, transferring from one language to another is insufficient for NLP systems to understand multilingual speakers in an increasingly multilingual world \citep{aronin2008multilingualism}. In many multilingual societies (e.g., Singapore, Papua New Guinea, etc.), it is common for multilingual interlocutors to produce sentences by mixing words, phrases, and even grammatical structures from the languages in their repertoires \citep{matras2007grammatical}. This is known as \emph{code-mixing} \citep{poplack1988social}, a phenomenon common in casual conversational environments such as social media and text messages.\footnote{Examples of real code-mixing in \Cref{app:examples}.} Hence, it is crucial for NLP systems serving multilingual communities to be robust to code-mixing if they are to understand and establish rapport with their users \citep{tay1989,bawa2020multilingual} or defend against adversarial polyglots.

%Existing work has presented datasets for evaluating code-mixed text processing ability \citep{bali2014borrowing,patwa2020sentimix}. While gold standard data is important for definitive evaluations, it is expensive to collect and annotate. The dizzying range of potential language combinations further compounds the immensity of such an effort.

Although gold standard data \citep{bali2014borrowing,patwa2020sentimix} is important for definitively evaluating code-mixed text processing ability, such datasets are expensive to collect and annotate. The dizzying range of potential language combinations further compounds the immensity of such an effort.

We posit that performance on appropriately crafted adversaries could act as a lower bound of a model's ability to generalize to the distribution simulated by said adversaries, an idea akin to \emph{worst-case analysis} \citep{divekar1984dc}.  For example, \citet{tan-etal-2020-bite} showed that an NLP system that was robust to morphological adversaries was less perplexed by dialectal text exhibiting morphological variation. Likewise, if a system is robust to code-mixed adversaries constructed from some set of languages, it is reasonable to expect it to also perform better on real code-mixed text in those languages. While they may not fully model the intricacies of real code-mixing \citep{sridhar1980syntax}, we believe that they can be useful in the absence of appropriate evaluation data. Hence, we:
\vspace{-0.5em}
\begin{itemize}[leftmargin=*]
\itemsep0em
    \item Propose two strong black-box adversarial attacks targeting the cross-lingual generalization ability of massive multilingual representations (\Cref{fig:bumblebee}), demonstrating their effectiveness on state-of-the-art models for natural language inference and question answering. To our knowledge, these are the first two multilingual adversarial attacks.
    \item Propose an efficient adversarial training scheme that takes the same number of steps as standard supervised training and show that it creates more language-invariant representations, improving accuracy in the absence of lexical overlap.
\end{itemize}

\section{Related Work}
%\vspace{-0.35em}
\paragraph{Multilingual classifiers.}
Low resource languages often lack support due to the high cost of annotating data for supervised learning. An approach to tackle this challenge is to build cross-lingual representations that only need to be trained on task data from a high resource language to perform well on another under-resourced language \citep{klementiev-etal-2012-inducing}. \citet{artetxe2019massively} presented the first general purpose multilingual representation using a BiLSTM encoder. Following the success of Transformer models \citep{vaswani2017attention}, recent multilingual models like mBERT \citep{bert19}, Unicoder \citep{huang2019unicoder}, and XLM-R \citep{conneau-etal-2020-unsupervised} take the pretraining$\rightarrow$fine-tuning paradigm into the multilingual realm by pretraining Transformer encoders on unlabeled monolingual corpora with various language modeling objectives before fine-tuning them on task data from a high-resource language such as English. This is known as cross-lingual transfer.

%\vspace{-0.5em}
\paragraph{Code-mixed text processing.}
Previous research on code-mixed text processing focused on constructing formal grammars \citep{joshi-1982-processing} and token-level language identification \citep{bali2014borrowing,solorio-etal-2014-overview,barman2014code}, before progressing to named entity recognition and part-of-speech tagging \citep{ball-garrette-2018-part,alghamdi2019leveraging,aguilar-solorio-2020-english}. Recent work explores code-mixing in higher-level tasks such as question answering and task-oriented dialogue \citep{chandu2019code,ahn2020code}.
\citet{muller2020can} demonstrate mBERT's ability to transfer to an unseen dialect by exploiting its speakers' tendency to code-mix.

A key challenge of developing models that are robust to code-mixing is the availability of code-mixed datasets. Hence, \citet{winata-etal-2019-code} use a pointer-generator network to generate synthetically code-mixed sentences while \citet{pratapa2018language} explore the use of parse trees for the same purpose.

\citet{yang2020code} propose to improve machine translation with ``code-switching pretraining", replacing words with their translations in a similar manner to masked language modeling \citep{bert19}. These word pairs are constructed from monolingual corpora using cosine similarity. \citet{sitaram2019survey} provide a comprehensive survey of code-mixed language processing.

\begin{table*}[t]
\small
    \centering
    \begin{tabular}{r  p{0.85\textwidth}}
    \toprule
    %\multicolumn{2}{c}{\textbf{XNLI}} \\
    %\midrule
    Original & \textbf{P:} The girl that can help me is all the way across town.
    \textbf{H:} There is no one who can help me. \\
    %Adversary & \textbf{P:} The girl that can help me is all the way across town. \textbf{H:} \begin{CJK*}{UTF8}{gbsn}没有人\end{CJK*} who can help me.\\
    Adversary & \textbf{P:} olan girl that can help me is all the way across town. \textbf{H:} \AR{لا يوجد} one who can help me. \\
    Prediction & \textbf{Before:} Contradiction \quad \textbf{After:} Entailment \\
    \midrule
    Original & \textbf{P:} We didn't know where they were going.  \textbf{H:} We didn't know where the people were traveling to.\\
    Adversary & \textbf{P:} We didn't know where they were going.  \textbf{H:} We didn't know where les gens allaient.\\
    Prediction & \textbf{Before:} Entailment \quad \textbf{After:} Neutral \\
    \midrule
    Original & \textbf{P:} Well it got to where there's two or three aircraft arrive in a week and I didn't know where they're flying to. \\\
    & \textbf{H:} There are never any aircraft arriving.\\
    Adversary & \textbf{P:} \foreignlanguage{russian}{общем, дошло до} mahali there's two or three aircraft arrive in a week and I didn't know where they're flying to. \\\
    & \textbf{H:} \begin{CJK*}{UTF8}{gbsn}从来没有\end{CJK*} aircraft arriving. \\
    Prediction & \textbf{Before:} Contradiction \quad \textbf{After:} Entailment \\
    \bottomrule
    \end{tabular}
    \vspace{-0.5em}
    \caption{\small \bee\ adversaries found for XLM-R on XNLI  (P: Premise; H: Hypothesis).}
    \label{tab:examples}
\vspace{-1em}
\end{table*}

%\vspace{-0.5em}
\paragraph{Word-level adversaries.}
Modified inputs aimed at disrupting a model's predictions are known as adversarial examples \citep{szegedy2014}.
In NLP, perturbations can be applied at the character, subword, word, phrase, or sentence levels.

Early word-level adversarial attacks \citep{ebrahimi-etal-2018-hotflip,blohm2018comparing} made use of the target model's gradients to flip individual words to trick the model into making the wrong prediction. However, while the perturbations were adversarial for the target model, perturbed word's original semantics was often not preserved. This could result in the expected prediction changing and making the model appear more brittle than it actually is.

Later research addressed this by searching for adversarial rules \citep{SinghGR18} or by constraining the candidate perturbations to the $k$ nearest neighbors in the embedding space \citep{alzantot-etal-2018-generating,michel-etal-2019-evaluation,ren2019generating,zhang-etal-2019-generating,li2019textbugger,jin2019bert}. \citet{zang2020word} take another approach by making use of a annotated sememes to disambiguate polysemous words, while \citet{tan-etal-2020-morphin} perturb only the words' morphology and encourage semantic preservation via a part-of-speech constraint. Other approaches make use of language models to generate candidate perturbations \citep{garg2020bae,han2020adversarial}. \citet{wallace-etal-2019-universal} find phrases that act as universally adversarial perturbations when prepended to clean inputs. \citet{zhang2019adversarial} provide a comprehensive survey.

%\vspace{-0.4em}
\paragraph{Summary.}
Existing work on pretrained multilingual models has highlighted their impressive zero-shot cross-lingual transfer ability, though some analyses \citep{K2020Cross-Lingual} indicate this could be a result of exploiting lexical overlaps rather than an indication of true cross-lingual understanding. Although language-agnosticity is commonly measured via cross-lingual retrieval tasks such as LAReQA \citep{roy-etal-2020-lareqa} and similarity search \citep{artetxe2019massively}, we offer a different perspective in this paper by operationalizing it as a model's ability to handle code-mixing. Existing evaluations for code-mixed text processing focus on gold annotated data, but such datasets are (relatively) expensive to compile and face similar scarcity challenges as those for low-resource languages. Existing word-/phrase-level adversarial attacks probing the limits of model robustness have largely focused on monolingual (English) inputs. In contrast, our adversarial attacks are designed to test the robustness of multilingual models to adversarial code-mixers. Finally, we propose an efficient adversarial training scheme to improve the robustness of said models to code-mixed adversaries.

\section{Generating Multilingual Adversaries}
%\vspace{-0.4em}
Code-mixing is a phenomenon where a multilingual speaker mixes words, and even grammatical rules, from different languages in a single sentence. This is distinguished from code-switching, which occurs at the inter-sentential level \citep{kachru1978mixing}.

%\vspace{-0.2em}
\paragraph{Extreme code-mixing.}
Inspired by the proliferation of real-life code-mixing and polyglots, we propose \poly\ and \bee, two multilingual adversarial attacks that adopt the persona of an adversarial code-mixer. We focus on the lexical component of code-mixing, where some words in a sentence are substituted with their equivalents from another language in the interlocutor's repertoire. Borrowed words fall into two categories, nonce borrowing and loanwords, though distinguishing between them is beyond the scope of this work.

Since most code-mixers are bilinguals, natural code-mixed sentences tend to be constructed from two languages, with one language determining the syntax of the overall sentence \citep{poplack1988social}. However, in a world with an increasing number of multilingual societies, it is conceivable for code-mixing to occur between more than two languages \citep{foley1988new}. We take this idea to the extreme to test multilingual representations for their robustness to such cross-lingual lexical variation.

\vspace{-0.2em}
\paragraph{Problem formulation.}
Given a target multilingual model $\gM$, a clean example $x$ with the label $y$, and a set of embedded languages $\sL$ from which to borrow words, we aim to generate the adversarial example $x'$ that maximizes $\gM$'s loss. Formally,
\begin{equation}\label{eq:loss}
x' = \underset{x_c \in X}{\argmax} \; \gL(y,\gM(x_c)),
\end{equation}
where $x_c \in X$ is a candidate adversary generated by perturbing $x$, $\gM$ is a task-specific neural model, and $\gL(\cdot)$ is the model's loss function.

\subsection{\poly: Word-Level Adversaries}
%vspace{-0.2em}
To obtain a code-mixed adversary, we first generate candidate adversaries by substituting words in the clean example with their equivalents from another language. These substitutions/perturbations can be generated by via machine translation or mined from bilingual dictionaries. Following \citet{myers1997duelling}, we will refer to the original example's language as the matrix language and the perturbation's language as the embedded language.

Next, we perform beam search on the candidates to find the adversary that maximizes the target model's loss in a black-box manner (\Cref{alg:polygloss} in \Cref{app:poly_impl}). In our implementation, we also keep track of successful adversaries and return the ones with the highest and lowest losses. The former is a stronger adversary, while the latter often has fewer perturbations. More details are in \Cref{app:poly_impl}.

%vspace{-0.2em}
\paragraph{Orthographic preservation.} When the embedded language uses a different script from the matrix language, code-mixers tend to transliterate borrowed words into the same script \citep{abuhakema2013code,bali2014borrowing}. This still poses a significant challenge to multilingual models \citep{khanuja-etal-2020-gluecos}. We generally preserve the embedded language's script where possible to avoid unfairly penalizing the target model since there is often no standard way of transliterating words.

%vspace{-0.2em}
\paragraph{Scalable sense disambiguation.}
Due to the polysemous nature of many words, translating the right sense is crucial to preserving the word's (and sentence's) semantics. Common word sense disambiguation methods \cite{wsd-eneko07} use a sense tagger trained on an annotated sense inventory such as WordNet \citep{Miller95wordnet}. However, this approach requires individual taggers and sense inventories for each matrix and embedded language, making it a serious challenge to extend \poly\ to low-resource languages.

Instead, we propose to filter candidate perturbations using the embedded language translation of the clean example. This is easily done by checking if the candidate perturbation exists in the translation. Since our examples tend to be single sentences, the probability of different senses of the same word occurring in a single sentence is generally low \citep{conneau-etal-2018-xnli,CUBBITT}. This approach only requires a machine translation (MT) system and no extra linguistic information, making it highly scalable as long as a supervised (or unsupervised) machine translation system is available. By using gold translations instead of machine translations, it is even possible to mostly guarantee semantic preservation at the word-level.

\begin{algorithm}[t]
\small
\begin{algorithmic}
\Require Clean example-label pair $(x, y)$, Target Model $\gM$,\\ Embedded languages $\sL$
\Ensure Adversarial example $x'$
%\State $S \gets \Call{Tokenize}{x}$
\State $T \gets \Call{Translate}{x, \text{target-languages} = \sL}$
\State $\gL_x \gets \Call{GetLoss}{\gM,x,y}$
\State $B \gets \{(\gL_x, x, 0)\}$
%\State $\sX' \gets \{\varnothing\}$
\algorithmiccomment{Initialize beam}
\State $\sP \gets \Call{AlignAndExtractPhrases}{x, T}$
\While{\Call{NotEmpty}{$B$}}
    \State ${\gL_{x_c}, x_c, i} \gets \Call{Poll}{B}$
    \State $C \gets \Call{GetCandidates}{x_c, \sP[i]}$
    \State $\gL \gets \Call{GetLoss}{\gM, C, y}$ \algorithmiccomment{Losses for $C$}
    %\If{}
    %\EndIf
    \State $i \gets i + 1$
    \State $\Call{UpdateBeam}{B, \gL, C, i}$
\EndWhile
\State $x' \gets \Call{Poll}{B}$
\State \Return $x'$
\caption{\bee}
\label{alg:bumblebee}
\end{algorithmic}
\end{algorithm}

\subsection{\bee: Phrase-Level Adversaries}
%vspace{-0.2em}
Although using bilingual dictionaries with our filtering method ensures that the semantics of a borrowed word matches the original, the dictionary's comprehensiveness determines the presence of sufficient candidate adversaries. In addition, \poly\ swaps words at the word level, which may hurt the naturalness of the resulting sentence since it is more common for code-mixers to borrow phrases than individual words \citep{abuhakema2013code}.

A solution to these issues is to replace phrases in the matrix sentence with their equivalents from the reference translations instead of using a dictionary lookup (\Cref{alg:bumblebee}). A key advantage of this approach is its flexibility and scalability to more languages since it only requires parallel bitexts from the matrix and embedded languages. With the advent of neural sequence-to-sequence models, such bitexts can be easily generated using publicly available MT models. However, a key challenge for this approach is extracting the matrix-embedded phrase pairs from the clean example and its translation. We follow common phrase-based machine translation methods and accomplish this by aligning the matrix and embedded sentences \cite{Koehn:2010:SMT}. Implementation details can be found in \Cref{app:bee_impl}.

%vspace{-0.2em}
\paragraph{Syntactic preservation.} \label{sec:syntax_preservation}
To improve the adversaries' naturalness, we impose an \emph{equivalence constraint} \citep{poplack1980sometimes}, preventing a perturbation from being applied if it is from the same language as the previous word \emph{and} will disrupt the syntax of the current phrase if applied \citep{winata-etal-2019-code}. Such disruptions usually occur when borrowing words from languages with a different word order.

\begin{table}[t]
\small%\fontsize{9}{11.5}
    \centering
    \begin{tabular}{l | c c c | c c c}
         \toprule
         %\vspace{0.1em}
         \textbf{\underline{Model}}  & \multicolumn{3}{c}{\underline{\smash{\textbf{XNLI-13}}}} & \multicolumn{2}{c}{\underline{\smash{\textbf{XNLI-31}}}}  \\
          & Clean & PG$_\text{uf.}$ & PG$_\text{filt.}$ & Clean & PG$_\text{filt.}$\\
         \midrule
         XLM-R$_\text{large}$ & 81.10 & \textbf{6.06} & 28.28 & 80.60 & \textbf{8.76 }\\
         XLM-R$_\text{base}$ & 75.42 & \textbf{2.17} & 12.27 & 74.75 & \textbf{3.57} \\
         mBERT$_\text{base}$ & 67.54 & \textbf{2.15} & 9.24 & 66.56 & \textbf{3.11} \\
         Unicoder$_\text{base}$ & 74.98 & \textbf{1.99} & 11.33 & 74.28 & \textbf{3.73} \\
         \bottomrule
    \end{tabular}
    %\vspace{-0.5em}
    \caption{\small \poly\ (PG) results (accuracy) on XNLI-13 and -31 test sets with beam width $= 1$.
    PG$_\text{\{filt., uf.\}}$ indicates whether the candidate perturbations were filtered using reference translations. Clean accuracy scores are the averages across all languages in the test set. Lower is better.}
    \label{tab:main_results_polygloss}
    %\vspace{-1em}
\end{table}

\begin{table}[t]
\small%\fontsize{9}{11.5}
    \centering
    \begin{tabular}{l | c c | c c c}
         \toprule
         %\vspace{0.1em}
         \textbf{\underline{Model}}  & \multicolumn{2}{c}{\underline{\smash{\textbf{XNLI-13}}}} & \multicolumn{3}{c}{\underline{\smash{\textbf{Standard XNLI}}}} \\
          & Clean & \textsc{Bb} & Clean & Rand. & \textsc{Bb} \\
         \midrule
         XLM-R$_\text{large}$ & 81.10 & \textbf{11.31} & 79.85 & 75.04 & \textbf{8.18} \\
         XLM-R$_\text{base}$ & 75.42 & \textbf{5.08} & 74.06 & 65.19 & \textbf{3.53}  \\
         mBERT$_\text{base}$ & 67.54 & \textbf{6.10} & 65.66 & 59.17 & \textbf{4.45} \\
         Unicoder$_\text{base}$ & 74.98 & \textbf{4.81} & 73.69 & 65.55 & \textbf{3.61} \\
         \bottomrule
    \end{tabular}
    %\vspace{-0.5em}
    \caption{\small \bee\ (\textsc{Bb}) results on XNLI with beam width $= 1$. \emph{Clean} accuracies are the averages across all languages in each test set. We include a \emph{random} (Rand.) baseline by randomly (rather than adversarially) perturbing sentences and report the average across 5 seeds. Lower is better.}
    \label{tab:main_results_bumblebee}
    %\vspace{-1em}
\end{table}

\iffalse
\begin{table*}[t]
\small%\fontsize{9}{11.5}
    \centering
    \begin{tabular}{l | c c c c | c c | c c c}
         \toprule
         %\vspace{0.1em}
         \textbf{\underline{Model}}  & \multicolumn{4}{c}{\underline{\smash{\textbf{XNLI-13 (Acc.)}}}} & \multicolumn{2}{c}{\underline{\smash{\textbf{Standard XNLI (Acc.)}}}} & \multicolumn{3}{c}{\underline{\smash{\textbf{XNLI-31 (Acc.)}}}}  \\
          & Clean & PG$_\text{unfilt.}$ & PG$_\text{filt.}$ & \bee\ & Clean & \bee\ & Clean & PG$_\text{unfilt.}$ & PG$_\text{filt.}$\\
         \midrule
         XLM-R$_\text{large}$ & 81.10 & 6.06 & 28.28 & 11.31 & 79.85 & 8.18 & 80.60 & 1.87 & 8.76 \\
         XLM-R$_\text{base}$ & 75.42 & 2.17 & 12.27 & 5.08 & 74.06 & 3.53 & 74.75 & 0.95 & 3.57 \\
         mBERT$_\text{base}$ & 67.54 & 2.15 & 9.24 & 6.10 & 65.66 & 4.45 & 66.56 & 0.91 & 3.11 \\
         Unicoder$_\text{base}$ & 74.98 & 1.99 & 11.33 & 4.81 & 73.69 & 3.61 & 74.28 & 0.87 & 3.73\\
         \bottomrule
    \end{tabular}
    \vspace{-0.5em}
    \caption{\small \poly\ (PG) and \bee\ results on various XNLI test sets with beam width $= 1$. % and alpha of 0.
    PG$_\text{\{fil., unfilt.\}}$ indicates whether the candidate perturbations were filtered using reference translations. Clean accuracy scores are the average scores across all languages in the test set. Lower is better.}
    \label{tab:main_results}
    \vspace{-1em}
\end{table*}
\fi

\section{Experiments}\label{sec:atk_exp}%vspace{-0.5em}
We first evaluate \poly\ and \bee\ on XNLI \citep{conneau-etal-2018-xnli}, then evaluate the stronger attack on XQuAD \citep{artetxe-etal-2020-cross}. XNLI is a multilingual dataset for natural language inference (NLI) with parallel translations for each example in fifteen languages. Each example comprises a premise, hypothesis, and a label with three possible classes: \{contradiction, neutral, entailment\}. We construct two more datasets from XNLI: XNLI-13 and XNLI-32. XNLI-13 comprises all XNLI languages except Swahili and Urdu due to the lack of suitable dictionaries for \poly. We then translate the English test set into eighteen other languages with MT systems to form XNLI-31, increasing the number of embedded languages \poly\ can draw from.
XQuAD is a multilingual dataset for extractive question answering (QA) with parallel translations in eleven languages. In the cross-lingual transfer setting, the models are trained on English data, MNLI \citep{mnli} and SQuAD 1.1 \citep{rajpurkar-etal-2016-squad}, and tested on mulitlingual data, XNLI and XQuAD, respectively. We perturb the premise and hypothesis for NLI and only the question for QA. More experimental details can be found in \Cref{app:experiment_details}.

%vspace{-0.2em}
\paragraph{Matrix language.} Although our attacks work with any language as the matrix language, we use English as the matrix language in our experiments due to the availability of English$\rightarrow$T translation models and the prevalence of English as the matrix language in many code-mixing societies.

%vspace{-0.2em}
\paragraph{Models.}
We conduct our experiments on three state-of-the-art massive multilingual encoder models: XLM-RoBERTa, mBERT, and Unicoder, each pretrained on more than 100 languages.

\begin{table}[t]
\small
    \centering
    \begin{tabular}{l | c c }
         \toprule
         \textbf{Model} & \textbf{Clean} & \textbf{\bee} \\
         \midrule
         XLM-R$_\text{large}$ & 75.64 / 61.39 & \textbf{35.32 / 22.52} \\
         XLM-R$_\text{base}$ & 68.90 / 53.50 & \textbf{17.95 / 10.33} \\
         mBERT$_\text{base}$ & 64.66 / 49.47 & \textbf{20.66 / 11.68} \\
         \bottomrule
    \end{tabular}
    \vspace{-0.5em}
    \caption{\small \bee\ results on XQuAD (F$_1$/EM).}
    \label{tab:xquad}
    \vspace{-1em}
\end{table}

\subsection{Results}
%vspace{-0.2em}
From \Cref{tab:main_results_polygloss,tab:main_results_bumblebee}, we observe that all the models are significantly challenged by adversarial code-mixing, though XLM-R$_\text{large}$ is the most robust to both attacks, likely due to having more parameters. However, even after filtering \poly's candidate perturbations by the gold translations in XNLI-13, we observe an average drop in accuracy of 80.01\%, relative to the models' accuracy on the clean XNLI-13. \bee\ induces even greater performance drops (average relative decrease of 90.96\% on XNLI-13), likely due to its word aligner yielding more candidates than \poly's dictionary lookup. Increasing the number of embedded languages \poly\ can draw upon results in greater drops in model performance (average relative decrease in accuracy of 93.66\% on XNLI-31).

%vspace{-0.25em}
\paragraph{BERT- vs. XLM-based.}
We notice that mBERT is more sensitive to intra-phrasal syntactic disruption than the XLM-based models. mBERT is the most robust to \bee\ out of all the base models when the equivalence constraint is in place, yet is the least robust to \poly. However, the latter trend is replicated for \bee\ if we remove this constraint (\Cref{tab:constraint_compare} in \Cref{app:extra_tables}). A \emph{possible} explanation is that XLM-R and Unicoder were trained on monolingual CommonCrawl (CC) data, while mBERT was trained on multilingual Wikipedia, which could be considered as aligned at the article level since there are articles on the same topic in different languages. Hence, it is possible that this helped to align the languages more accurately in the feature space but made it more sensitive to syntactic disruptions. However, many other hyperparameters differ between the two that could have also influenced their robustness. Hence, we leave a rigorous study of these factors to future work. The higher performance of the XLM-based models on clean data can likely be attributed to the CC corpus being an order of magnitude larger than multilingual Wikipedia \citep{lauscher2020zero}.

%vspace{-0.25em}
\paragraph{Candidate filtering.}
In the \emph{unfiltered} setting, it is impossible for \poly\ to discriminate between valid and invalid senses for a given context. Hence, a potential criticism is that the large difference in \poly's success rate between the filtered and unfiltered settings could be attributed to the inappropriate senses of polysemous words being chosen and disrupting the semantics of the sentence. On the other hand, filtering perturbations with reference translations of the sentence shrinks the space of perturbations to \mytilde 1 per language. Due to the dictionaries' non-exhaustive nature, not every word in the matrix sentence has an entry in the dictionary to begin with, making this filtering step a significant reduction of the space of candidates.

To determine the likely cause of the accuracy difference between the filtered and unfiltered settings in XNLI-13, we increase the number of languages available to \poly\ to thirty-one. If the difference between the filtered and unfiltered settings were \emph{not} due to a lack of sufficient candidates, we should observe only a minor difference between the filtered settings for both XNLI-13 and -31. However, we observe a 69\% drop for XLM-R$_\text{large}$, indicating that the former accuracy difference is likely due to the reduced number of valid candidates.

%vspace{-0.2em}
\paragraph{Phrase-level adversaries.}
In addition to generating more fluent sentences (\Cref{tab:examples}), extracting the candidate perturbations directly from the translations does away with the need for sense disambiguation and increases the number of perturbations per example since it is not limited to a static dictionary. The increased effectiveness of \bee\ compared to \poly\ (1.13x) is further evidence that a key factor to the success of such adversarial attacks is the availability of sufficient candidates; increasing the dimensionality of the search space increases the probability that an adversarial example for the model exists \citep{goodfellow2015}. We also include a non-adversarial baseline (Rand.) by sampling candidates from a uniform distribution instead of searching for the worst-case perturbations. Our results in \Cref{tab:main_results_bumblebee} indicate that the worst-case performance of multilingual models on code-mixed data may be much lower than the scores reported on human-produced test sets since they were not created in a targeted, adversarial fashion.
Experiments on beam width and a proof of concept for fully unsupervised adversaries are in \Cref{app:extra_exp}.

\begin{table}[t]
\small%\fontsize{9}{11.5}
    \centering
    \begin{tabular}{l | c | c }
         \toprule
         %\vspace{0.1em}
         \textbf{Model} & \textbf{Devanagari} & \textbf{Transliterated (Latin)}  \\
         \midrule
         XLM-R$_\text{large}$ & 61.35 & \textbf{41.97}\\
         XLM-R$_\text{base}$ & 48.62 & \textbf{30.01}\\
         mBERT$_\text{base}$ & 37.70 & \textbf{23.41}\\
         Unicoder$_\text{base}$ & 49.34 & \textbf{30.00}\\
         \bottomrule
    \end{tabular}
    %\vspace{-0.5em}
    \caption{\bee\ results on XNLI$_{\text{en,hi}}$ using both Devanagari and Latin scripts. Lower is better.}
    \label{tab:translit}
    %\vspace{-1em}
\end{table}

\paragraph{Transliteration.}
Since real-life code-mixers often use a single script for the entire sentence, we now test the effect of transliteration on \bee's success rate for the English + Hindi language pair. We accomplish this by transliterating all candidates from Devanagari into Latin using the dictionaries released by \citet{roark-etal-2020-processing}.
From \Cref{tab:translit}, we see that transliteration significantly affects the robustness of all models, even the XLM-based ones which were pretrained on similar data.

%vspace{-0.2em}
\paragraph{XQuAD.}
We observe that both XLM-R and mBERT are significantly challenged by \bee\ even though only the question was modified (\Cref{tab:xquad}). We did not experiment on Unicoder to reduce carbon costs since its performance was almost identical to XLM-R$_\text{base}$ in our XNLI experiments.

%vspace{-0.2em}
\paragraph{\poly\ or \bee?}
As expected, inspection of individual adversarial examples revealed that \bee\ generated more natural sentences than \poly\ since the languages used within phrases were more consistent (\Cref{tab:examples}). However, incorrect alignments due to the word aligner's probabilistic nature could introduce occasional noise into the adversarial examples. For example, we found ``the" (en) to be often aligned with ``\begin{CJK*}{UTF8}{gbsn}的\end{CJK*}" (zh) even though the former is an article and the latter a possessive. We observe that the aligner performs better when the sentences have similar word orders (e.g., English-French vs. English-Chinese) and we can expect the adversaries generated in these settings to be more natural. Hence, we recommend \poly\ when greater preservation of word-level semantics is desired, and \bee\ when phrase-level perturbations are desired or bilingual dictionaries are unavailable.

%vspace{-0.2em}
\paragraph{Discussion.}
\citet{K2020Cross-Lingual} noted significant performance drops in XNLI accuracy for mBERT when the premise and hypothesis were in different languages (Fake English vs. \{Hindi, Russian, Spanish\}), theorizing this to be an effect of disrupting the model's reliance on lexical overlap. Our experiments in \Cref{sec:atk_exp} and \Cref{sec:cat} lend support to this hypothesis. In \Cref{tab:examples}, we see multiple examples where the prediction was flipped from ``contradiction" to ``entailment" simply by perturbing a few words. If the models did not rely on lexical overlap but performed comparisons at the semantic level, such perturbations should not have severely impacted their performance. Our results on QA also corroborate \citet{lee-etal-2019-latent}'s finding that models trained on SQuAD-style datasets exploit lexical overlap between the question and context.

\begin{table*}[t]
\small
    \centering
    \begin{tabular}{l | c c c | c c c | c c}
         \toprule
         \textbf{Condition/Method} & \textbf{Clean} & \textbf{Clean$_\text{DL}$} &  \textbf{Adv$_\text{\Cref{sec:atk_exp}}$} &  \textbf{Adv$_\text{sw}$} & \textbf{Adv$_\text{hi+ur}$} & \textbf{Adv$_\text{XNLI}$}  & \textbf{Adv$_\text{tl}$} & \textbf{Adv$_\text{id+tl}$}  \\
         \midrule
         Cross-lingual transfer (from \Cref{sec:atk_exp}) & 74.06 & 66.09 & 3.53 & 38.54 & 29.12 & 3.53 & 36.96 & 24.83 \\
         Translate-train-$n$ & \textbf{77.25} & 72.01 & 29.44 & 50.53 & 40.63 & 7.04 & 44.37 & 33.23 \\
         DANN \citep{ganin2016domain} & 51.86 & 35.10 & 33.45 & 16.02 & 17.52 & 6.54 & 12.05 & 7.06\\
         Code-mixed adv. training (CAT) & 77.10 & \textbf{75.46} & \textbf{50.21} & \textbf{58.58} & \textbf{48.20} & \textbf{12.63} & \textbf{49.14} & \textbf{38.16}\\
         \bottomrule
    \end{tabular}
    \vspace{-0.5em}
    \caption{\small Results on standard XNLI with XLM-R$_\text{base}$. Clean refers to the combined test set of all languages, Clean$_\text{DL}$ to the variant where the hypothesis and premise of each example are from different languages, Adv$_\text{\Cref{sec:atk_exp}}$ to the \bee\ adversaries from \Cref{sec:atk_exp}, and Adv$_\text{\{lgs\}}$ to new adversaries from English + the subscripted languages. Higher is better.}
    \label{tab:adv_train}
    \vspace{-1em}
\end{table*}

\section{Code-Mixed Adversarial Training}
\label{sec:cat}
%vspace{-0.3em}
Finally, we propose code-mixed adversarial training (CAT), an extension of the standard adversarial training paradigm \citep{goodfellow2015}, to improve the robustness of multilingual models to adversarial polyglots. In standard adversarial training, adversarial attacks are run on the training set to generate adversaries for training. However, this makes adversarial training computationally expensive. Hence, we take inspiration from \citet{tan-etal-2020-morphin}'s method of randomly sampling perturbations from an adversarial distribution and generate code-mixed perturbations using word alignment.

To generate the code-mixed adversarial training set $X'$, we first compute the adversarial distribution $\gP_{adv}$ by enumerating the perturbations per embedded language in all successful adversaries (\Cref{sec:atk_exp}). Formally, $\gP_{adv}=\{f_i\}_{i=1...|\sL|},$ where $f_i = \frac{l_i}{\sum^{|\sL|}_{j=1} l_j}$ and $\sL$ is the set of embedded languages.

Next, for each clean example $x$, we sample $n$ languages from $\gP_{adv}$ before translating the example into the $n$ languages and aligning the translations with $x$. For sentence-pair classification tasks like NLI, we use a per-sentence $n$ to further increase variation. Intuitively, limiting $n$ improves the example's naturalness and the algorithm's efficiency (the alignment is the most costly step). We then extract phrases from the aligned sentences, yielding our candidate perturbations $\sP$. Next, we sample a perturbation with probability $\rho$ from $\sP$ for each phrase in $x$. Reducing $\rho$ yields more natural sentences since they will be less perturbed. Finally, we apply these perturbations to $x$, obtaining a CAT example $x'$. Doing this $k$ times for all $x$ in $X$ and adding the result to $X$ yields $X'$ (\Cref{alg:codemixer} in \Cref{app:extra_tables}).

In contrast to running the adversarial attack on the training set, sampling perturbations from a distribution does not guarantee that the resulting example will be adversarial to the model. This issue can be mitigated by increasing the number of CAT examples observed during training. However, this would increase the computational cost if we were to train the model for the same number of epochs. Hence, we set $k$ to one less than the number of epochs XLM-R$_\text{base}$ was fine-tuned for in \Cref{sec:atk_exp} and train the model for one epoch on the adversarial training set. This exposes the model to more variation in the same number of training steps.

%vspace{-0.2em}
\paragraph{Setting.}
We conduct our experiments on NLI with XLM-R$_\text{base}$ with no loss of generality. In \Cref{sec:atk_exp}, the model was trained for ten epochs. Hence, we set $k=9, n=2, \rho=0.5$ for CAT and train all models for a similar number of steps (60k) with the same hyperparameters as \Cref{sec:atk_exp}. We first test the models on the \bee\ adversaries generated in \Cref{sec:atk_exp} before directly attacking the model. Next, we construct more realistic settings by running \bee\ with only 1-2 embedded languages from standard XNLI, Swahili (sw), Hindi (hi), and Urdu (ur). These languages were the lowest resourced in the pretraining data \citep{conneau-etal-2020-unsupervised}.

We also construct another \emph{non-adversarial} test set from XNLI by randomly choosing hypotheses and premises from different languages \citep{K2020Cross-Lingual}. Since the original examples are individually monolingual, this test set will reveal if a model is simply exploiting lexical overlap rather than comparing the underlying concepts.

Finally, we run \bee\ with embedded languages not seen during task-specific training \emph{and} from a different family (Austronesian) from the XNLI languages, Filipino (tl) and Indonesian (id). This zero-shot defense setting will reveal if CAT encourages the learning of more language-invariant representations, or is simply allowing the model to adapt to the adversarial distribution.

%vspace{-0.5em}
\paragraph{Baselines.}
Since training on languages in the test set takes us out of the cross-lingual transfer setting, we train a \emph{translate-train-$n$} baseline for a fair comparison. In this setting, we train on every $x$ and its translations in the $n$ languages sampled in CAT, regardless of whether they contributed words to the final CAT examples. We also include \citet{ganin2016domain}'s domain adversarial neural network (DANN), which has been used for cross-lingual adaptation \citep{joty-etal-2017-cross,chen2018adversarial}.

\subsection{Results}
%vspace{-0.2em}
From \Cref{tab:adv_train}, we observe that both training on fully translated data and on CAT examples improved accuracy on the non-adversarial test sets and robustness to code-mixed adversaries, compared to the cross-lingual transfer model that was only trained on English data. Similar to \citet{K2020Cross-Lingual}, we found that disrupting the models' reliance on lexical overlap (Clean$_\text{DL}$) hurt performance. The drop was particularly significant for the cross-lingual transfer (8 points) and \emph{translate-train-$n$} models (5.24 points). On the other hand, our CAT model only suffered a 1.5-point drop, indicating that the former two models likely rely heavily on lexical overlap to make predictions, while our CAT model may be using ``deeper", more language-agnostic features. Crucially, our CAT model achieves similar to better clean accuracy than the baselines, contrasting with prior work showing that adversarial training hurts clean accuracy \citep{tsipras2018robustness}.
Finally, our CAT model is $>$1.7x more robust to adversaries constructed from all fifteen XNLI languages than the \emph{translate-train-$n$} model. Although DANN-type training improved robustness to the previous \bee\ adversaries, clean performance was significantly degraded and \bee\ was able to find even more damaging adversaries upon attacking the model directly.

When attacked with 1-2 embedded languages that were seen during training, CAT also yields significant improvements in robustness over the baselines: a $>7$ point increase compared to \emph{translate-train-$n$} and a $>$19 point gain over the zero-shot transfer setting. In the zero-shot defense setting, CAT shows a $>$12-point gain over the zero-shot transfer model and a $>$4.7-point gain over the \emph{translate-train-$n$} model. We believe these results to be due to CAT encouraging the learning of language-invariant representations by exposing the model to cross-lingual lexical variation and preventing the model from exploiting lexical overlaps.

\section{Seeing is Believing}
%vspace{-0.5em}
\begin{figure}[t]
\vspace{-0.25em}
\small
    \centering
    \begin{subfigure}{0.237\textwidth}
    \includegraphics[width=\textwidth]{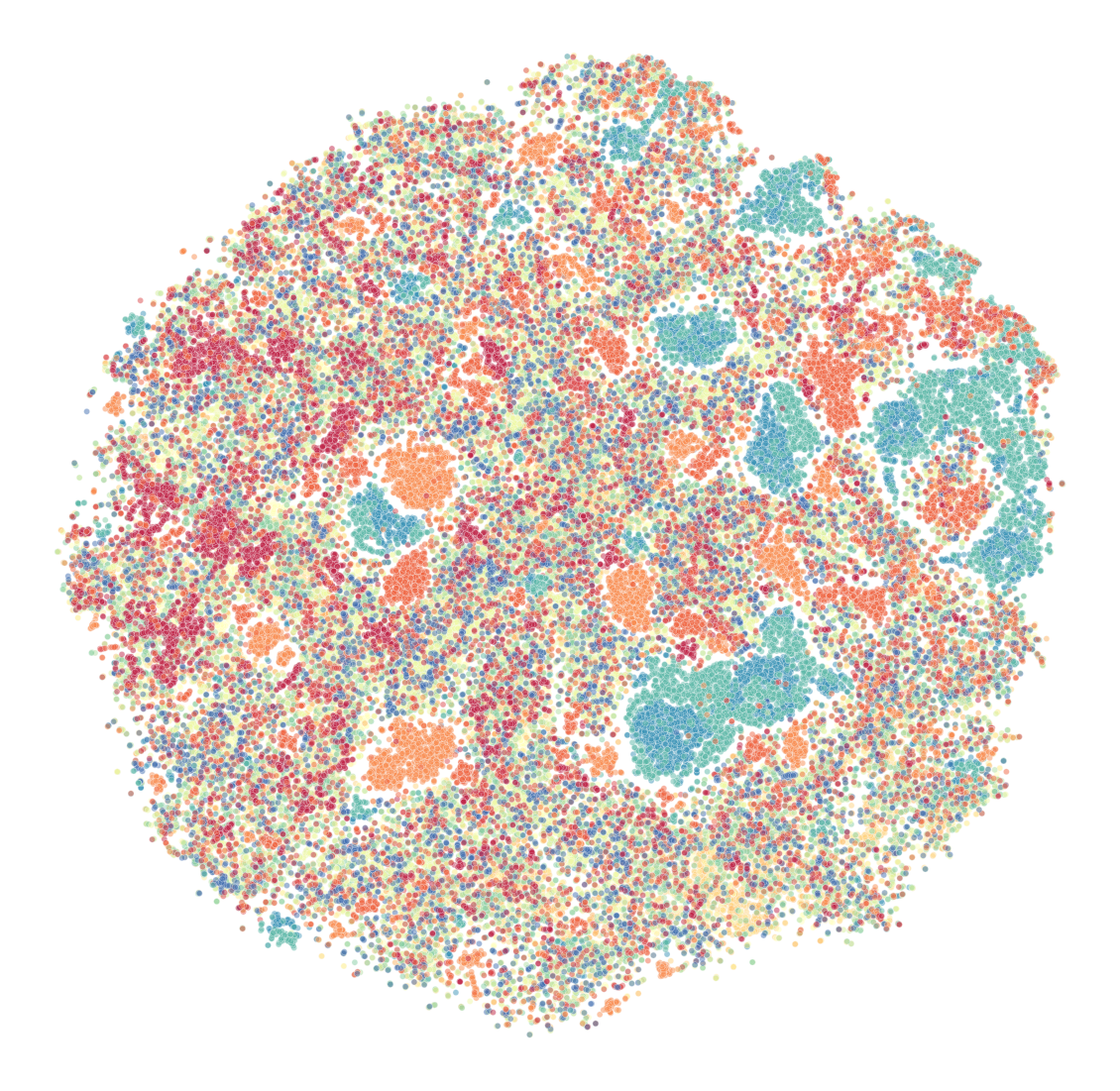}
    %\vspace{-0.25em}
    \caption{Cross-Lingual Transfer}
     \label{fig:xling}
    \end{subfigure}
    \hfill
    \begin{subfigure}{0.237\textwidth}
    \includegraphics[width=\textwidth]{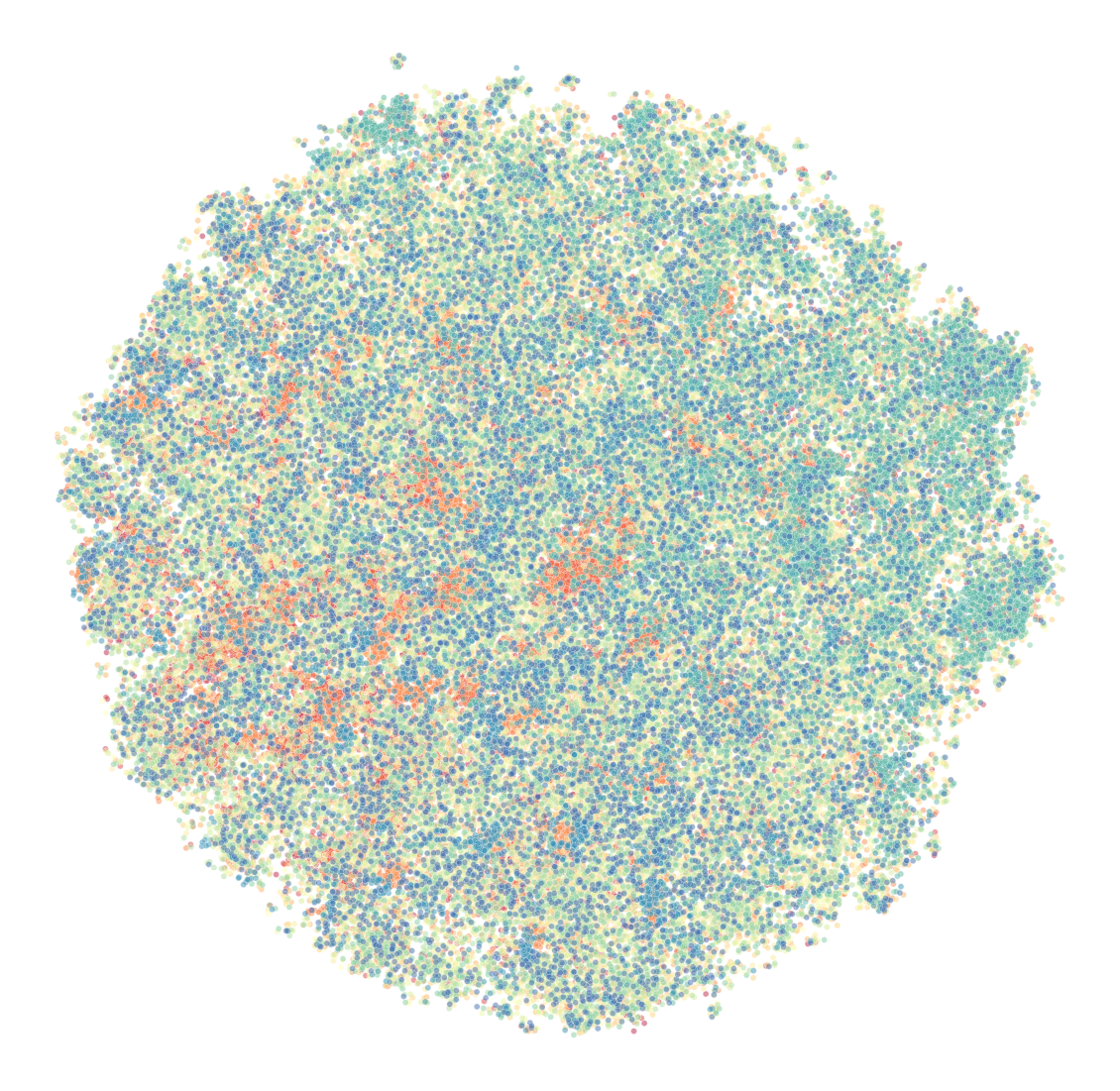}
    %\vspace{-0.25em}
    \caption{CAT}
     \label{fig:cat}
    \end{subfigure}
    \caption{t-SNE visualizations of XLM-R$_\text{base}$ representations fine-tuned using different methods.}
    \label{fig:tsne_main}
    \vspace{-1em}
\end{figure}
To further understand the effect of various fine-tuning methods on XLM-R$_\text{base}$, we visualize the \texttt{<s>} vector from the layer before the classification head using t-SNE \citep{linderman2019fast}. Here, all sentences from XNLI are passed through the representations individually. If a representation were 100\% language-invariant, we should expect t-SNE to be unable to separate individual languages into their own clusters. Hence, the extent to which t-SNE is able to do so would indicate the amount of language-specific information in this last layer.

From \Cref{fig:xling}, we observe that for the cross-lingual transfer model (\Cref{sec:atk_exp}), t-SNE managed to organize the sentences from several languages (Chinese, Hindi, Thai, Urdu) into distinct clusters. This indicates that a significant amount of language-specific information remains in the vector representations of sentences from these languages. Visualizing the sequence-averaged embeddings makes this even clearer (\Cref{fig:tsne_avg_app} in \Cref{app:extra_tables}). Hence, while XLM-R may be multilingual, it appears to be structured as a space of individual language subspaces as opposed to a mixed, or language-invariant space. On the other hand, t-SNE was much less successful when given the representation trained with CAT (\Cref{fig:cat}). Mixing multiple languages in the same sentence and showing the model multiple variants of the same sentence likely encourages the model to refine its representation such that all variants of the same sentence are represented similarly, resulting in a more language-invariant representation. T-SNE plots of the other models are in \Cref{app:extra_tables}.

\section{Limitations and Future Work}
%vspace{-0.75em}
We acknowledge that our methods do not fully model real code-mixing since we do not learn the mixing patterns from real data and there are subtleties in real code-mixing we ignore for simplicity, e.g., accounting for the prestige of participating languages \citep{bhatia2011multilingual}. In addition, it is impossible to \emph{guarantee} the semantic preservation of a sentence generated by \bee\ due to the word aligner's statistical nature, though we can expect more accurate alignments to improve semantic preservation.  Finally, while CAT improves robustness, there remains a significant gap between the robust and clean accuracies. In line with recent work challenging the Anglocentricity of cross-lingual models \citep{anastasopoulos-neubig-2020-cross,Liang2020XGLUEAN}, a promising direction of future work lies in investigating how the choice of matrix language affects model robustness.

\section{Conclusion}
%vspace{-0.75em}
Ensuring that multilingual models are robust to both natural and adversarial code-mixing is important in today's increasingly multilingual world if they are to allow their target users to fully express themselves in human-machine conversations and to defend against adversarial users attempting to evade toxicity/misinformation detection systems.

To approximate a lower bound for model performance on lexical code-mixing, we propose two strong black-box multilingual adversarial attacks and demonstrate their effectiveness on state-of-the-art cross-lingual NLI and QA models. The former generates perturbations from bilingual dictionaries and disambiguates between senses using sentence translations, while the latter generates perturbations by aligning sentences from different languages.

Next, we show that training on code-mixed data synthesized via word alignment improves clean and robust accuracy when models are prevented from exploiting lexical overlap \emph{without} hurting clean accuracy. Crucially, we achieve this in the same number of steps as standard supervised training.

Finally, we use t-SNE visualizations to show that multilingual models are not necessarily language-invariant and that our code-mixed adversarial training scheme encourages language-invariance.

\section{\hspace{-0.53em}Broader Impact / Ethical Considerations}
%vspace{-0.5em}
Adversarial attacks and defenses are double-edged swords. On one hand, adversarial examples expose the gaps in existing models and help to focus the research community's attention on flaws that need to be addressed before these models can be used reliably in noisy, real-world environments. On the other, the same adversarial attacks can be used by malicious actors to bypass toxicity/misinformation detection systems. Similarly, methods for improving adversarial robustness can be used to defend against malicious actors and improve robustness to natural noise or linguistic variation, yet they can also be used to strengthen automated censorship systems and limit freedom of speech. For example, our adversarial attacks could be used both as a lower bound for model performance on naturally occurring code-mixed text \emph{and} to bypass misinformation detection systems while preserving the message's intelligibility for multilingual speakers. Our adversarial training method could be used to both improve machine understanding of code-mixers by making multilingual representations more language-invariant \emph{and} suppress the freedom of speech of polyglots who could have been using code-mixing to evade censorship.

At the same time, technology strongly shapes our behavior \citep{reeves2019screenomics}. Consequently, given the centrality of code-switching/mixing to many polyglots' lived experiences \citep{duff2015transnationalism} and the positive correlations between multilingualism, code-switching, and creativity \citep{leikin2013effect,kharkhurin2015role,furst2018multilingualism}, we should ensure that the natural language technologies we build do not inhibit multilingual speakers from fully expressing themselves, e.g., by discouraging code-mixing due to non-understanding. In addition, studies have found that aphasic polyglots code-mix more frequently than neurotypical polyglots to cope with word-retrieval difficulties \citep{goral2019variation}, making it important for natural language technologies to be robust to code-mixing if they are to be inclusive. Therefore, we include both adversary generation and defense methods to avoid tipping the balance too far in either direction.

\section*{Acknowledgments}
We would like to thank Cynthia Siew (NUS Psychology), Greg Bennett and Kathy Baxter (Salesforce), Lav Varshney (UIUC Electrical and Computer Engineering), Min-Yen Kan (NUS Computer Science), and our anonymous reviewers for their invaluable feedback. We are also grateful to Guangsen Wang, Mathieu Ravaut, Soujanya Lanka, and Tuyen Hoang for contributing manually code-mixed sentences and Bari M Saiful for pointers on replicating the XLM-R results. Samson is supported by Salesforce and Singapore's Economic Development Board under its Industrial Postgraduate Programme.

\bibliography{bibs/fairness,bibs/nlp,bibs/ling,bibs/multilingual,bibs/adversarial,bibs/misc}
\bibliographystyle{acl_natbib}

\clearpage
\appendix
\renewcommand*{\thesubsection}{\Alph{section}.\arabic{subsection}}
\section{Examples of Real Code-Mixed Text}
\label{app:examples}
\textbf{English + Spanish} \citep{sridhar1980syntax}
\vspace{-0.5em}
\begin{itemize}
\itemsep0em
    \item The man \hl{que vino ayer} (who came yesterday) wants \hl{ito buyun carro nuevo} (a new car).
    \item \hl{El} (The) old man \hl{esta enojide} (is mad).
    \item Me lleve chile ya \hl{roasted} y \hl{peeled} ... para hacerlo ells. (I picked up the chile already roasted and peeled for making it there.)
\end{itemize}

\noindent\textbf{Hindi + English}
\citep{bali2014borrowing}
\vspace{-0.5em}
\begin{itemize}
\itemsep0em
    \item Befitting reply to \hl{mere papa ne maaraa} (My father gave a befitting reply)
    \item ... and  the  party  workers  [will]  come  with  me without \hl{virodh} (protest/objection)
\end{itemize}

\noindent\textbf{Sarnami + Sranan + Dutch}
\citep{yakpo2015code}
\vspace{-0.5em}
\begin{itemize}
\itemsep0em
    \item {\sethlcolor{cyan}\hl{dus gewoon}} calat jaiye, tab ego \hl{kerki} {\sethlcolor{cyan}\hl{links}} ki {\sethlcolor{cyan}\hl{rechts}}. (So just keep on walking, then [there’s] achurch, left or right.)
    \item kaun wálá  \underline{d}am\underline{r}ú, ego haigá jaun me\underline{n} ná \hl{verfi} bhail, ma ego {\sethlcolor{cyan}\hl{wel}} hai. (Which [kind of] damru drum, there’s one which is not coloured inside but one actually is.)
\end{itemize}

\section{Attack Implementation Details}
\subsection{\poly}
\label{app:poly_impl}
\begin{algorithm}[h]
\small
\begin{algorithmic}
\Require Clean example-label pair $(x, y)$, Target Model $\gM$,\\ Embedded languages $\sL$
\Ensure Adversarial example $x'$
%\State $S \gets \Call{Tokenize}{x}$
\State $T \gets \Call{Translate}{x, \text{target-languages} = \sL}$
\State $\gL_x \gets \Call{GetLoss}{\gM,x,y}$
\State $B \gets \{(\gL_x, x, 0)\}$ \algorithmiccomment{Initialize beam}
\While{\Call{NotEmpty}{$B$}}
    \State ${\gL_{x_c}, x_c, i} \gets \Call{Poll}{B}$
    \State $C \gets \Call{GetCandidates}{x_c, \text{token-id} =i}$
    \State $C \gets \Call{FilterCandidates}{C, T}$
    \State $\gL \gets \Call{GetLoss}{\gM, C, y}$ \algorithmiccomment{Losses for $C$}
    \State $i \gets i + 1$
    \State $\Call{UpdateBeam}{B, \gL, C, i}$
\EndWhile
\State $x' \gets \Call{Poll}{B}$
\State \Return $x'$
\caption{\poly}
\label{alg:polygloss}
\end{algorithmic}
\end{algorithm}
To reduce the practical running time of our attack, we make use of cross-lingual dictionaries released by \citet{lample2018muse} for generating candidate perturbations instead of translating the words in an online fashion. We also use the gold translations of the clean examples when they are available (such as in XNLI), and use the models released by \citet{TiedemannThottingal:EAMT2020} in the \texttt{transformers} library \citep{Wolf2019HuggingFacesTS} to translate the examples to other languages. We also cache them in hashtables for fast retrieval.

\subsection{\bee}
\label{app:bee_impl}
We use the gold translations where available and \citet{TiedemannThottingal:EAMT2020}'s translation models for the other languages, and align sentences with a neural word aligner \citep{sabet2020simalign} backed by XLM-R$_\text{base}$ in our implementation of \bee. Although \citet{sabet2020simalign} found the ``Itermax" algorithm to yield the best performance for their experimental settings, we suggest using the high recall (``Match") algorithm for candidate generation. We inspected the output of both algorithms and found that while Itermax generates more candidates, it also tends to generate noisier alignments compared to Match, which we found to be more conservative.

\section{\poly\ and CAT Samples}
\vspace{-0.5em}
Almost random samples for \poly\ and CAT (we tried to include the sentences with Thai and Hindi characters but did not manage to get them to render with pdflatex).
\begin{table}[h]
\small
    \centering
    \begin{tabular}{p{\linewidth}}
    \toprule
   Rockefeller {\greektext έδ}$\upomega${\greektext σε} to \AR{السرطان} forschung.\\
   \midrule
   	The \begin{CJK*}{UTF8}{gbsn}发音 列表\end{CJK*} \foreignlanguage{russian}{само} included the most basic things.\\
    \midrule
    He vardı yedi hermanas and no brothers in his family.\\
    \midrule
    But même \foreignlanguage{russian}{хотя} as a boy I lived on a çiftlik right on the Mexican \AR{حدود}, I \AR{تذكر} being mystified by ranching términos that crept into Western songs du north of us, cayuse, for example.\\
    \midrule
	We should nie think of human equality when we consider social and political justice.\\
    \bottomrule
    \end{tabular}
    \caption{\poly\ adversaries for XLM-R$_\text{base}$ on XNLI-13.}
    \label{tab:poly_samples}
\end{table}

\begin{table}[h]
\small
    \centering
    \begin{tabular}{p{\linewidth}}
    \toprule
    Not much has sich innerhalb des Jahrzehnts verändert.\\
    \midrule
    Ese \AR{كان} el most historic weather \AR{يوم} in la historia registrada para el el clima severo en el country \\
    \midrule
    Como ustedes {\greektext γν}$\upomega${\greektext ρίζετε}, last enero hemos issued un {\greektext νέο όγκο} de reports, la Performance {\greektext και λογοδοσίας} serie de, en los que se esbozan los de gestión desafíos {\greektext που αντιμετ}$\upomega${\greektext πίζουν οι} nuestras mayores federal agencies and the substantial opportunities para mejorar su rendimiento.\\
    \midrule
    because a lot \foreignlanguage{vietnamese}{trong số họ} are are similar	\\
    \midrule
    um-hum bueno that's increíble como i used to cuando i was \foreignlanguage{russian}{в} college solía la have el stereo en all \foreignlanguage{russian}{през} time or i tenía en MTV or something but ever que i've been out de college\\
    \bottomrule
    \end{tabular}
    \caption{Code-mixed adversarial training examples.}
    \label{tab:cat_samples}
\end{table}

\section{Experiment Details}
\label{app:experiment_details}
\subsection{Datasets}
Standard XNLI\footnote{\href{https://cims.nyu.edu/~sbowman/xnli}{cims.nyu.edu/\mytilde sbowman/xnli}} comprises 7,500 \emph{parallel} examples (2,490 dev., 5,010 test) in fifteen languages: English (en), Spanish (es), German (de), Greek (el), Russian (ru), Turkish (tr), Arabic (ar), Vietnamese (vi), Thai (th), Chinese (zh), Hindi (hi), French (fr), Bulgarian (bg), Swahili (sw), and Urdu (ur). The labels are uniformly distributed between the three classes (contradiction, neutral, entailment).

The machine-translated training set\footnote{Same link as XNLI dev/test set.} of standard XNLI comprises the MNLI training examples in addition to their translations in the same fourteen non-English languages as the dev. and test sets.

XNLI-13 comprises all standard XNLI languages except Swahili and Urdu due to the lack of suitable dictionaries for \poly. XNLI-31 comprises all languages in XNLI-13, in addition to another eighteen: Afrikaans (af), Albanian (sq), Catalan (ca), Czech (cs), Danish (da), Dutch (nl), Estonian (et), Filipino (tl), Finnish (fi), Hebrew (he), Hungarian (hu), Indonesian (id), Italian (it), Macedonian (mk), Romanian (ro), Slovak (sk), Swedish (sv), and Ukrainian (uk).

MNLI\footnote{\href{https://cims.nyu.edu/~sbowman/multinli}{cims.nyu.edu/~sbowman/multinli/}} comprises 392,702 examples in English with the following label distribution: 130,899 entailment, 130,900 neutral, 130,903 contradiction.

XQuAD\footnote{\href{https://github.com/deepmind/xquad}{github.com/deepmind/xquad}} comprises 1,190 question-answer pairs with guaranteed answers in eleven languages: English (en), Spanish (es), German (de), Greek (el), Russian (ru), Turkish (tr), Arabic (ar), Vietnamese (vi), Thai (th), Chinese (zh), and Hindi (hi). XQuAD examples are drawn from the SQuAD 1.1 development set.

SQuAD 1.1\footnote{\href{https://rajpurkar.github.io/SQuAD-explorer/dataset/train-v1.1.json}{rajpurkar.github.io/.../train-v1.1.json}} comprises 87,599 question-answer pairs with guaranteed answers in English.

\subsection{Metrics}
The metric used for XNLI is simple accuracy:
\begin{equation}
   \text{Accuracy} = \frac{\# \, \text{true positive}}{\# \, \text{total}}
\end{equation}

\noindent
The metrics used for SQuAD are F$_1$:
\begin{equation}
\text{F}_1 = \frac{2 \cdot \text{precision} \cdot \text{recall}}{\text{precision} + \text{recall}}
\end{equation}
and exact match:
\begin{equation}
\text{Exact Match} =\frac{\# \, \hat{y}=y}{\# \, \text{total}}
\end{equation}

\subsection{XQuAD Preprocessing}
\label{app:xquad}
We use \citet{pythainlp} to tokenize Thai text, \texttt{jieba}\footnote{\href{https://github.com/fxsjy/jieba}{github.com/fxsjy/jieba}} for Chinese text, and split all other languages on whitespace.

\subsection{Training Details}
\label{app:training_details}
\begin{table}[H]
\small
    \centering
    \begin{tabular}{l c c c c}
         \toprule
         \textbf{Model} & \textbf{Params.} & \textbf{Lr} & \textbf{Bsz} & \textbf{Epochs}\\
         \midrule
         XLM-R$_\text{large}$ & 550M & 1e-06 & 64 & 10 \\
         XLM-R$_\text{base}$ & 270M & 5e-06 & 64 & 10 \\
         mBERT$_\text{base}$ & 172M & 5e-05 & 64 & 2 \\
         Unicoder$_\text{base}$ & 270M & 5e-06 & 64 & 10 \\
         \bottomrule
    \end{tabular}
    \caption{Hyperparameters for model fine-tuning on XNLI (MNLI). Number of parameters reproduced from \citet{conneau-etal-2020-unsupervised}.}
    \label{tab:xnli_hparams}
\end{table}

\begin{table}[H]
\small
    \centering
    \begin{tabular}{l c c c}
         \toprule
         \textbf{Model} & \textbf{Learning rate} & \textbf{Batch Size} & \textbf{Epochs} \\
         \midrule
         XLM-R$_\text{large}$ & 1e-05 & 32 & 3 \\
         XLM-R$_\text{base}$ & 3e-05 & 64 & 2 \\
         \bottomrule
    \end{tabular}
    \caption{Hyperparameters for model fine-tuning on XQuaD (SQuAD 1.1).}
    \label{tab:xquad_hparams}
\end{table}

\paragraph{Hyperparameters.}
\Cref{tab:xnli_hparams} contains the hyperparameters we used to fine-tune our models on MNLI. We used the hyperparameters suggested by \citet{bert19}\footnote{\href{https://github.com/google-research/bert/blob/master/multilingual.md}{https://github.com/google-research/.../multilingual.md}} for mBERT, the hyperparameters suggested by \citet{bari2020multimix} for the XLM-R models, and the hyperparameters from \citet{Liang2020XGLUEAN} for Unicoder.

\Cref{tab:xquad_hparams} contains the hyperparameters we used to fine-tune our models on SQuAD 1.1. We used the default SQuAD hyperparameters from the \texttt{transformers} library\footnote{\href{https://github.com/huggingface/transformers/tree/master/examples/question-answering}{github.com/huggingface/.../question-answering}} for XLM-R$_\text{base}$ and adjusted the hyperparameters for XLM-R$_\text{large}$ to fit it onto the GPU. The mBERT model from the HuggingFace model repository\footnote{\href{https://huggingface.co/salti/bert-base-multilingual-cased-finetuned-squad}{huggingface.co/salti/bert-base-multilingual-cased-finetuned-squad}} was used instead of fine-tuning our own. The clean scores reported in \Cref{tab:xquad} are similar to those reported in the XQuAD GitHub repository.

\subsection{Validation Performance for CAT Exps.}
\begin{table}[H]
\small
    \centering
    \begin{tabular}{l c c}
         \toprule
         \textbf{Model} & \textbf{Clean} & \textbf{Adv$_\text{\Cref{sec:atk_exp}-dev}$} \\
         \midrule
          Cross-lingual transfer (from \Cref{sec:atk_exp}) & 73.90 & 4.17\\
         Translate-train-$n$ & 76.99 & 29.27  \\
         DANN \citep{ganin2016domain} & 51.81 & 34.89 \\
         Code-mixed adv. training (CAT) & \textbf{77.13} & \textbf{52.32} \\
         \bottomrule
    \end{tabular}
    \caption{Accuracy on the XNLI dev.\ set and \bee\ adversaries generated from the dev.\ set.}
    \label{tab:dev}
\end{table}

\subsection{Infrastructure Details}
Models were trained on single V100 GPUs. Attacks were run on 8 V100 GPUs, parallelized with \texttt{ray}\footnote{\href{https://github.com/ray-project/ray}{github.com/ray-project/ray}} to make full use of GPU memory. On the standard XNLI test set (15 languages, 5,010 examples), \bee\ takes 60-90 minutes in total. With 1 embedded language, \bee\ runs on the test set in under 10 minutes. \poly\ is generally much faster (under 30 minutes on XNLI-31) due to not needing a neural aligner.

\section{Extra \bee\ Experiments}
\label{app:extra_exp}
\begin{table}[h]
\small
    \centering
    \begin{tabular}{l c c c}
         \toprule
         \textbf{Variable ($v$)} & \textbf{$v=1$} &  \textbf{$v=2$} & \textbf{$v=3$} \\
         \midrule
         Beam width & 48.52\% & 48.95\% & 49.67\% \\
         Embedded lgs. & 48.52\% & 69.09\% & 75.73\%  \\
         \bottomrule
    \end{tabular}
    \caption{Effect of increasing the beam width vs. the number of embedded languages on the attack success rate (\%) while holding the other variable constant at 1. We use Swahili, Swahili and French, and Swahili, French, and Spanish as the embedded languages when $v=1,2,3$, respectively. Rates are computed relative to the average clean XNLI score on the languages involved.}
    \label{tab:beam}
\end{table}

\paragraph{Beam search.}
In our experiments, we found that increasing the beam width yielded a higher attack success rate. However, this increases running time with only minor improvements (\Cref{tab:beam}). We found increasing the number of embedded languages (and hence candidates) to be a more efficient method of increasing the success rate with a minor increase in running time. Although the time complexity (in number of model queries) is $O(|B||C||\sL||S|)$ where $|S|$ is the sentence length, increasing $|\sL|$ had a greater impact on the success rate than increasing $|B|$ by the same number.

\begin{table}[h]
\small
    \centering
    \begin{tabular}{l c c c }
         \toprule
         \textbf{Model} & \textbf{Clean} & \textbf{Supervised} &  \textbf{Unsupervised} \\
         \midrule
          XLM-R$_\text{base}$ & 81.32 & 49.00 & 46.78 \\
         \bottomrule
    \end{tabular}
    \caption{Accuracy after running \bee\ on XNLI in supervised and unsupervised settings (matrix: English, embedded: French).}
    \label{tab:unsup}
\end{table}
\paragraph{Fully unsupervised adversaries.}
A potential drawback of \bee\ is that it requires translations of the clean example, which may be challenging to obtain for low-resource languages. However, it is possible to use unsupervised MT models for this purpose. We use \citet{song2019mass}'s unsupervised English-French model to generate translations as a proof of concept and find that they achieve similar results (\Cref{tab:unsup}).

\section{Plausible Language Combinations}
To explore the effect of different combinations of languages on a multilingual model's performance, we run \bee\ with different sets of embedded languages that could be plausibly spoken by (adversarial) polyglots around the world. English is used as the matrix language for all experiments.

We observe a general trend of decreasing model accuracy as we increase the number of mixed languages (\Cref{tab:multilgs}), and XLM-R$_\text{base}$'s robustness to an embedded language appears to be generally more dependent on the size of its pretraining dataset than language typology. For example, Russian (\textit{en+ru}) and Indonesian (\textit{en+id}), languages with notably high accuracies, were the two most resourced languages after English in the corpus used for pretraining \citep{conneau-etal-2020-unsupervised}, while Swahili and Filipino were among the lower-resourced languages. A notable outlier is Afrikaans (\textit{en+af}), which was the lowest resourced language in our experiments yet XLM-R$_\text{base}$ was quite robust to adversaries constructed from it and English. A possible explanation is that Afrikaans' language family, Indo-European, is highly represented in the pretraining corpus. Another notable outlier is Vietnamese, which was the fourth most resourced language in the pretraining corpus, yet the model was more vulnerable to adversaries constructed from English and Vietnamese than adversaries constructed from English and Filipino, one of the lowest resourced languages. A \emph{possible} explanation for this is the use of Latin characters combined with little vocabulary overlap between English and Vietnamese, and differing adjective positions.

\begin{table}[h]
\small
    \centering
    \begin{tabular}{c c c c}
         \toprule
        %size of resource pair?
       %distribution of codemixing, say like code mixing but its not
       %someone who knows multiple languages might try to attack the model this way, but not necessarily what they naturally do
        \textbf{Lgs. (en+)} & \textbf{Exemplar Region} &  \textbf{Acc.} & \textbf{Size (GB)} \\
         \midrule
         tr & Turkey & 55.76 & 20.9\\
         de & Germany & 55.42 & 66.6\\
         bg & Bulgaria & 54.87 & 57.5\\
         ru & Russia & 54.33 & 278.0 \\
         th & Thailand & 52.43 & 71.7 \\
         el & Greece & 52.41 & 46.9\\
         zh & China & 52.17 & 46.9 \\
        ar & Middle East & 51.53 & 28.0 \\
         es & Spain & 50.07 & 53.3\\
         fr & France & 49.00 & 56.8 \\
         hi & India & 48.62 & 20.2 \\
         ur & Pakistan & 42.09 & 5.7\\
         sw & Kenya & 38.54 & 1.6 \\
         vi & Vietnam & 36.32 & 137.3 \\
         \midrule
         ro & Romania & 56.26 & 61.4 \\

         id & Indonesia & 50.91 & 148.3 \\
        
         af & Namibia & 49.30 & 1.3 \\
        
        sq & Albania & 40.13 & 5.4 \\
        
        tl & Philippines & 36.96 & 3.1 \\
        
         \midrule
         ru+uk & Ukraine & 44.37 & 362.6 \\
        ar+he & Israel & 37.24 & 59.6 \\
        af+de & Namibia & 36.46 & 67.9 \\
        de+fr & Switzerland & 35.84 & 123.4 \\
        id+zh & Indonesia & 35.30 & 195.2\\
        hu+ro & Transylvania & 34.11 & 119.8 \\
        ar+fr & Morocco & 32.15 & 84.8\\
        hi+ur & Kashmir & 29.12 & 25.9\\
        ar+sw & Tanzania & 24.37 & 29.6 \\
        \midrule
        fi+sv+ru	& Finland & 30.27 & 344.4 \\
        cs+hu+sk & Slovak & 30.21 & 97.2\\
        es+fr+it & S. Europe & 26.58 & 140.3\\
        mk+sq+tr & Albania & 23.53 & 31.1\\
        \midrule
        da+de+nl+sv & Denmark & 27.74 & 153.6 \\
        id+th+tl+vi & S.E. Asia & 12.53 & 360.4\\
         \bottomrule
    \end{tabular}
    \caption{How XLM-R$_\text{base}$ \emph{might} fare against real-life adversarial polyglots and where they might be found. Scores are accuracy on the XNLI test set with English as the matrix language. Corpora sizes reproduced from \citet{conneau-etal-2020-unsupervised}.}
    \label{tab:multilgs}
\end{table}

We also find XLM-R$_\text{base}$ to be twice as robust in the \textit{en+da+de+nl+sv} condition as the \textit{en+id+th+tl+vi} condition. It is likely that the structural similarity of the mixed languages also plays an important role in determining the model robustness, reinforcing \citet{K2020Cross-Lingual}'s similar findings on cross-lingual transfer. Investigating the effects of typology, vocabulary and orthographic overlap, pretraining corpus size, and interaction effects between different sets of languages on a model's robustness to code-mixed adversaries could lead to new insights into how different languages interact in the multilingual embedding space and we leave this to future work.

\begin{table*}[]
\small
    \centering
    \begin{tabular}{l | p{0.8\textwidth}}
         \toprule
         English &
\textbf{P:} Americans should also consider how to do it-organizing their government in a different way.\\
& \textbf{H:} The American government might be organized in a different way.	 \\
\cmidrule{2-2}
French (R) & \textbf{P:} Les Américains devraient aussi réfléchir to de le faire, en organisant in la different way.	\\
& \textbf{H:} The gouvernement américain might be organized in a different way.\\
\cmidrule{2-2}
French (\textsc{Bb}) & \textbf{P:} Americans should also penser à comment organiser leur gouvernement differemment.\\
& \textbf{H:} Le gouvernement americain est maybe organized differently.\\
\cmidrule{2-2}
Hindi (R) & \textbf{P:} Americans ko yeh bhi sochna hai ki kaise karna hai-organizing their government in a different way. \\
& \textbf{H:} America ki sarkar might be organized in a different way. \\
\cmidrule{2-2}
Hindi (\textsc{Bb}) & \textbf{P:} amerikiyon chahiye ki also vichar kese to karna it-organizing their government in a different way.\\
& \textbf{H:} The American government might be organized in a different way. \\
\cmidrule{2-2}
Chinese (R) & \textbf{P:} Americans\begin{CJK*}{UTF8}{gbsn}应该\end{CJK*}consider\begin{CJK*}{UTF8}{gbsn}下怎样去重组他们的\end{CJK*}government\\
& \textbf{H:} \begin{CJK*}{UTF8}{gbsn}美国\end{CJK*}government\begin{CJK*}{UTF8}{gbsn}可以有其它的\end{CJK*}organization\begin{CJK*}{UTF8}{gbsn}方式\end{CJK*}\\
\cmidrule{2-2}
Vietnamese (R) & \textbf{P:} \foreignlanguage{vietnamese}{Người Mỹ cũng nên} consider how to do it-organizing their government \foreignlanguage{vietnamese}{theo một cách khác}.\\
& \textbf{H:} \foreignlanguage{vietnamese}{Chính phủ Mỹ có thể được} organized \foreignlanguage{vietnamese}{theo một cách khác}.\\
\midrule
English &
\textbf{P:} When that occurs, the lending fund sacrifices interest from Treasury securities on its invested balances and instead receives interest from the borrowing fund on the amount of the loan.	\\
& \textbf{H:} The lending fund doesn't get all the interest in some cases.\\
\cmidrule{2-2}
French (R) & \textbf{P:} Lorsque cela se produit, le lending fonds de sacrifie interest des Trésor securities sur ses investis balances et plutôt receives intérêts du borrowing fund sur le montant du les loan. \\
& \textbf{H:} The lending fund ne reçoit pas all the interest in some cases.\\
\cmidrule{2-2}
French (\textsc{Bb}) & \textbf{P:} Quand ça arrive, le lending fund sacrifie l'intéret des Treasury securities sur ses investissements et recoit des interest from the borrowing fund on the amount du prêt.\\
& \textbf{H:} Le fending fund ne reçoit pas tous les interest in some cases.\\
\cmidrule{2-2}
Hindi (R) & \textbf{P:} Jab aisa hota hai, the lending fund sacrifices interest from Treasury securities on its invested balances and instead receives interest from the borrowing fund on the amount of the loan.\\
& \textbf{H:} Aise mamlo mein, the lending fund doesn’t get all the interest.\\
\cmidrule{2-2}
Chinese (R) & \textbf{P:} \begin{CJK*}{UTF8}{gbsn}这种情况下，\end{CJK*}lending fund\begin{CJK*}{UTF8}{gbsn}会损失从\end{CJK*}treasury securities\begin{CJK*}{UTF8}{gbsn}的\end{CJK*}investment interests\begin{CJK*}{UTF8}{gbsn}，但是可以从\end{CJK*}borrowing fund\begin{CJK*}{UTF8}{gbsn}那里拿到\end{CJK*}interest\\
& \textbf{H:} lending fund\begin{CJK*}{UTF8}{gbsn}在某些\end{CJK*}case\begin{CJK*}{UTF8}{gbsn}下不会拿到所有的\end{CJK*}interest\\
\cmidrule{2-2}
Chinese (\textsc{Bb}) & \textbf{P:} \begin{CJK*}{UTF8}{gbsn}当\end{CJK*} that occurs, the lending fund sacrifices \begin{CJK*}{UTF8}{gbsn}利息 从\end{CJK*} Treasury securities \begin{CJK*}{UTF8}{gbsn}在 其 投资 余额 ，而\end{CJK*} receives \begin{CJK*}{UTF8}{gbsn}利息\end{CJK*} from the borrowing \begin{CJK*}{UTF8}{gbsn}基金\end{CJK*} on the \begin{CJK*}{UTF8}{gbsn}金额 的\end{CJK*} the loan. \\
& \textbf{H:} The lending \begin{CJK*}{UTF8}{gbsn}基金并\end{CJK*} doesn't get all the interest in some cases.\\
\cmidrule{2-2}
Vietnamese (R) & \textbf{P:} \foreignlanguage{vietnamese}{Khi điều ấy xảy ra}, the lending fund sacrifices interest from Treasury securities on its invested balances \foreignlanguage{vietnamese}{và thay vào đó} receives interest from the borrowing fund on the amount of the loan.\\
& \textbf{H:} The lending fund doesn't get all the interest \foreignlanguage{vietnamese}{trong một số trường hợp}.\\
\cmidrule{2-2}
Vietnamese (\textsc{Bb}) & \textbf{P:} \foreignlanguage{vietnamese}{Khi đó} occurs, \foreignlanguage{vietnamese}{điều quỹ cho vay hy suất từ chứng khoán mình} invested \foreignlanguage{vietnamese}{số dư và thay đó nhận được lãi} borrowing \foreignlanguage{vietnamese}{tiền trên} the \foreignlanguage{vietnamese}{số} of \foreignlanguage{vietnamese}{tiền} loan.\\
& \textbf{H:} The lending fund doesn't get all \foreignlanguage{vietnamese}{cả sự quan tâm} some cases.\\
\midrule
English &
\textbf{P:} It's the truth, you fool. \quad \textbf{H:} Everything I say is true.\\
\cmidrule{2-2}
French (R) & \textbf{P:} It's the truth, you fool. \quad \textbf{H:} Tout ce que je true.\\
\cmidrule{2-2}
French (\textsc{Bb}) & \textbf{P:} C'est la truth, abruti. \quad \textbf{H:} Tout ce que je dis is true. \\
\cmidrule{2-2}
Hindi (R) & \textbf{P:} Yaha sach hai, you fool. \quad \textbf{H:} Everything I say, sach hai.\\
\cmidrule{2-2}
Hindi (\textsc{Bb}) & \textbf{P:} It's the truth, you fool. \quad \textbf{H:} jo I say is true.\\
\cmidrule{2-2}
Chinese (R) & \textbf{P:} \begin{CJK*}{UTF8}{gbsn}傻瓜，这就是\end{CJK*}truth \quad \textbf{H:} \begin{CJK*}{UTF8}{gbsn}我说的都是\end{CJK*}truth\\
\cmidrule{2-2}
Chinese (\textsc{Bb}) & \textbf{P:} \begin{CJK*}{UTF8}{gbsn}这是\end{CJK*} the truth, you fool. \quad \textbf{H:} \begin{CJK*}{UTF8}{gbsn}我说的一切\end{CJK*} true.\\
\cmidrule{2-2}
Vietnamese (R) & \textbf{P:} \foreignlanguage{vietnamese}{Đó là sự thật}, you fool. \quad \textbf{H:} Everything I say \foreignlanguage{vietnamese}{là đúng}.\\
\cmidrule{2-2}
Vietnamese (\textsc{Bb}) & \textbf{P:} It's the truth, you fool. \quad \textbf{H:} \foreignlanguage{vietnamese}{Tất} I say is true.\\
\bottomrule
    \end{tabular}
    \caption{A comparison of code-mixed examples produced by bilinguals (R) and \bee\ (\textsc{Bb}). Since the humans were only provided with the English examples, some differences in phrasing is to be expected.}
    \label{tab:qual_samples}
\end{table*}

\newpage

\section{More Tables, Figures, and Algorithms}
\label{app:extra_tables}
\begin{algorithm}[h]
\small
\begin{algorithmic}
\Require Original examples $X$, Embedded languages $\sL$, \\ Num. perturbed examples $k$, Adversarial distribution $\gP_{adv}$, \\ Max. langs. per example $n$, Phrase perturbation prob. $\rho$

\Ensure Adversarial training set $X'$
\State $X' \gets \{\varnothing\} $
\For {$x$ in $X$}
\State $\sS \gets \Call{SampleLanguages}{\sL, n, \gP_{adv}}$
\State $T \gets \Call{Translate}{x, \text{target-languages} = \sS}$
\State $\sP \gets \Call{AlignAndExtractPhrases}{x, T}$
\For {$i = 1$ to $k$}
    \State $x' \gets \Call{Perturb}{x, \sP, \rho}$
   
\State $X' \gets X' \cup \{x'\}$
\EndFor
\EndFor
\State \Return $X \cup X'$
\caption{CAT Example Generation}
\label{alg:codemixer}
\end{algorithmic}
\end{algorithm}

\begin{figure}[h]
    \centering
    \includegraphics[width=0.5\textwidth]{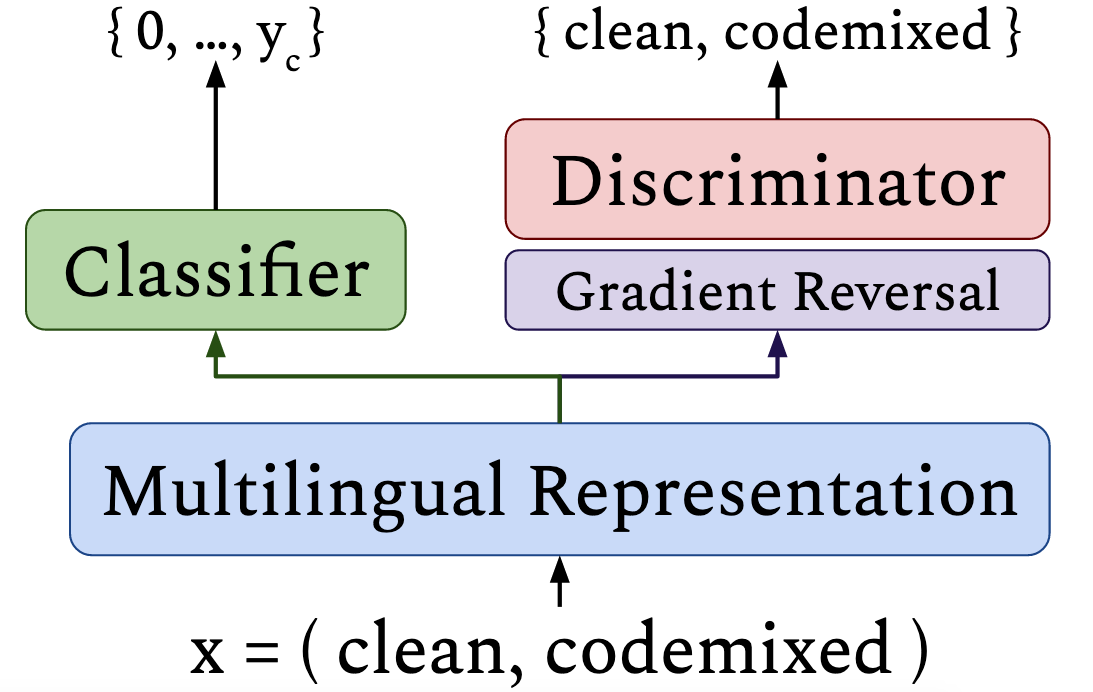}
    \caption{Domain Adversarial Neural Network}
    \label{fig:dann}
\end{figure}
Section continues on next page.
\begin{table*}[h]
\fontsize{9}{11.5}
    \centering
    \begin{tabular}{c | c c c c c}
         \toprule
         \textbf{Model}
          & \textbf{Clean} & \textbf{PG$_\text{unfilt.}$} & \textbf{PG$_\text{filt.}$} & \textbf{\bee} & \textbf{\bee$_\text{unconstr}$}\\
         \midrule
         XLM-R$_\text{large}$ & 81.10 & 6.06 & 28.28 & 11.31 & 5.22 \\
         XLM-R$_\text{base}$ & 75.42 & 2.17 & 12.27 & 5.08 & 1.47 \\
         mBERT$_\text{base}$ & 67.54 & 2.15 & 9.24 & 6.10 & 1.19 \\
         Unicoder$_\text{base}$ & 74.98 & 1.99 & 11.33 & 4.81 & 1.29 \\
         \bottomrule
    \end{tabular}
    \caption{\poly\ (PG) and \bee\ results on the XNLI-13 test set with a beam width of 1. 
    PG$_\text{\{fil., unfilt.\}}$ indicates whether the candidate substitutions were filtered using reference translations. \bee$_\text{unconstr}$ refers to the setting without the equivalence constraint (\Cref{sec:syntax_preservation}).}
    \label{tab:constraint_compare}
\end{table*}

\begin{figure*}[h]
    \small
    \centering
    \begin{subfigure}{0.49\textwidth}
    \includegraphics[width=\textwidth]{figs/zeroshot_tsne_xnli_cls_sb_no_legend.png}
    \caption{Cross-Lingual Transfer}
    \end{subfigure}
    \hfill
    \begin{subfigure}{0.49\textwidth}
    \includegraphics[width=\textwidth]{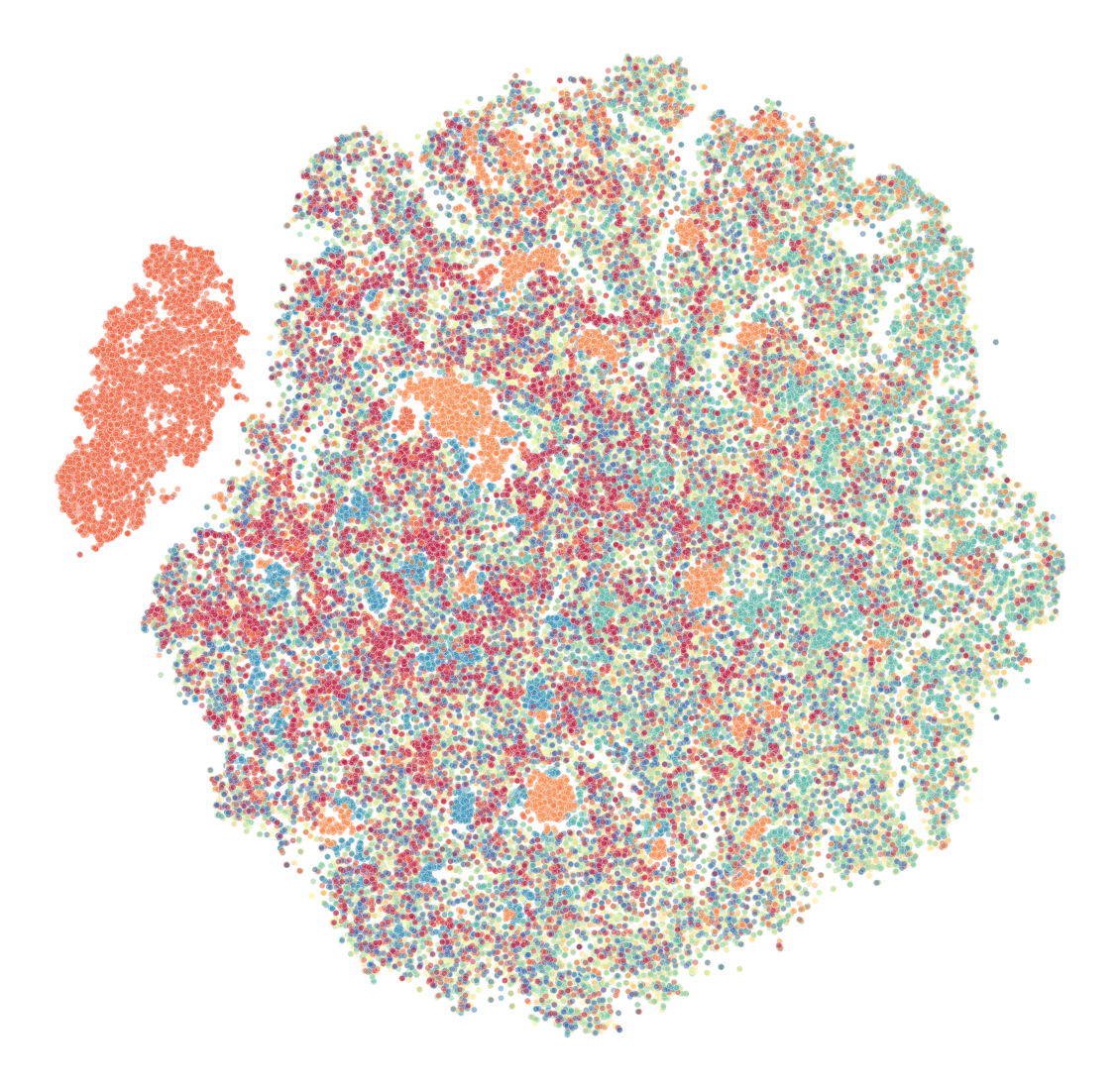}
    \caption{Translate-Train-$n$}
    \end{subfigure}
    \begin{subfigure}{0.49\textwidth}
    \includegraphics[width=\textwidth]{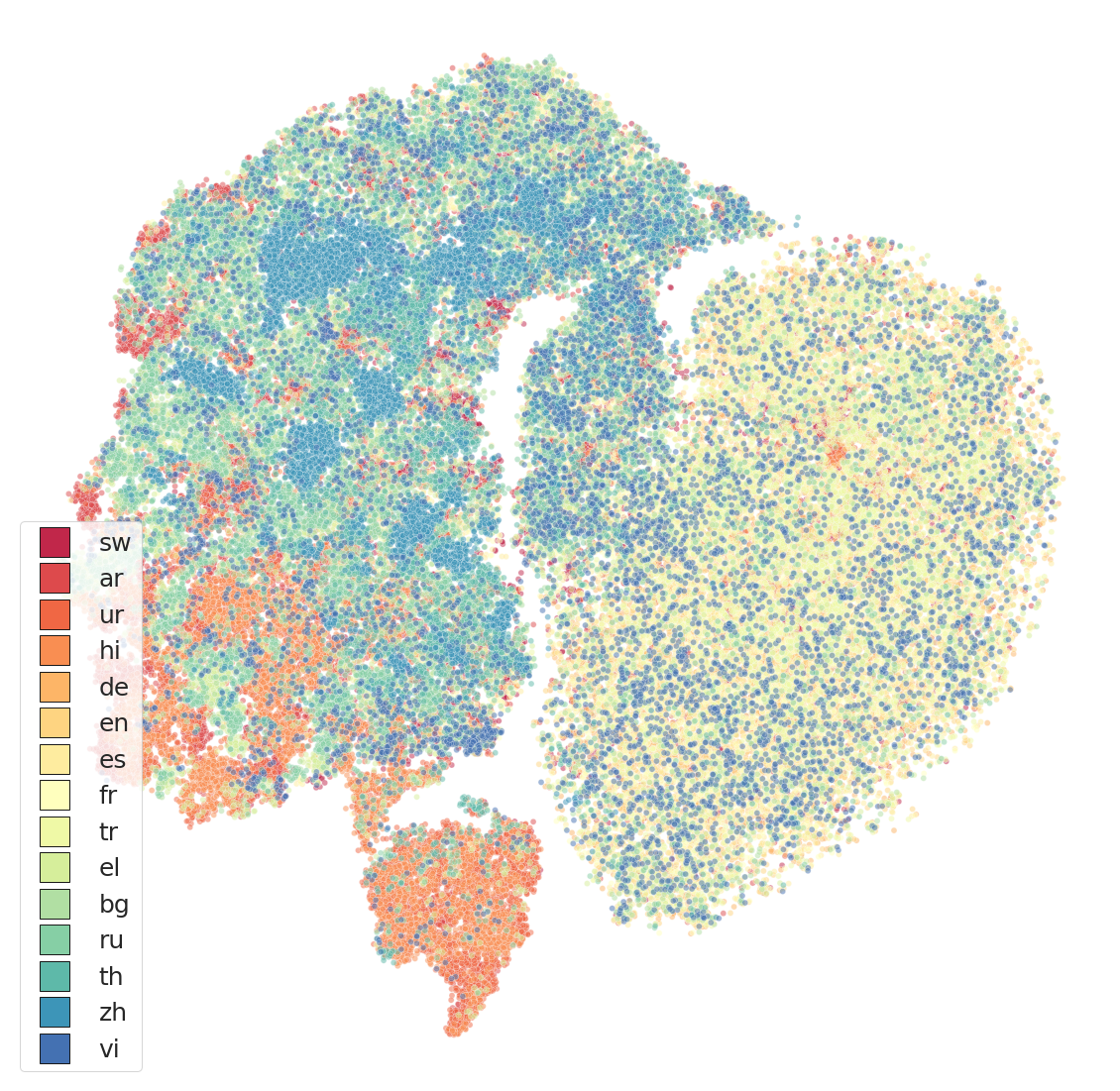}
    \caption{Domain Adversarial Neural Network}
    \label{fig:tsne_dann}
    \end{subfigure}
    \hfill
    \begin{subfigure}{0.49\textwidth}
    \includegraphics[width=\textwidth]{figs/cat_tsne_xnli_cls_sb_no_legend.png}
    \caption{Code-Mixed Adversarial Training}
    \end{subfigure}
    \caption{t-SNE visualizations of XLM-R$_\text{base}$ representations fine-tuned using different methods. We use \citet{linderman2019fast}'s t-SNE algorithm implemented in openTSNE \citep{Policar731877} and tried to arrange related languages close to each other on the color spectrum (to the extent possible on one dimension) so it would be obvious if similar languages were getting clustered together, as in (c).}
    \label{fig:tsne_app}
\end{figure*}

\begin{figure*}[h]
    \centering
    \begin{subfigure}{0.49\textwidth}
    \includegraphics[width=\textwidth]{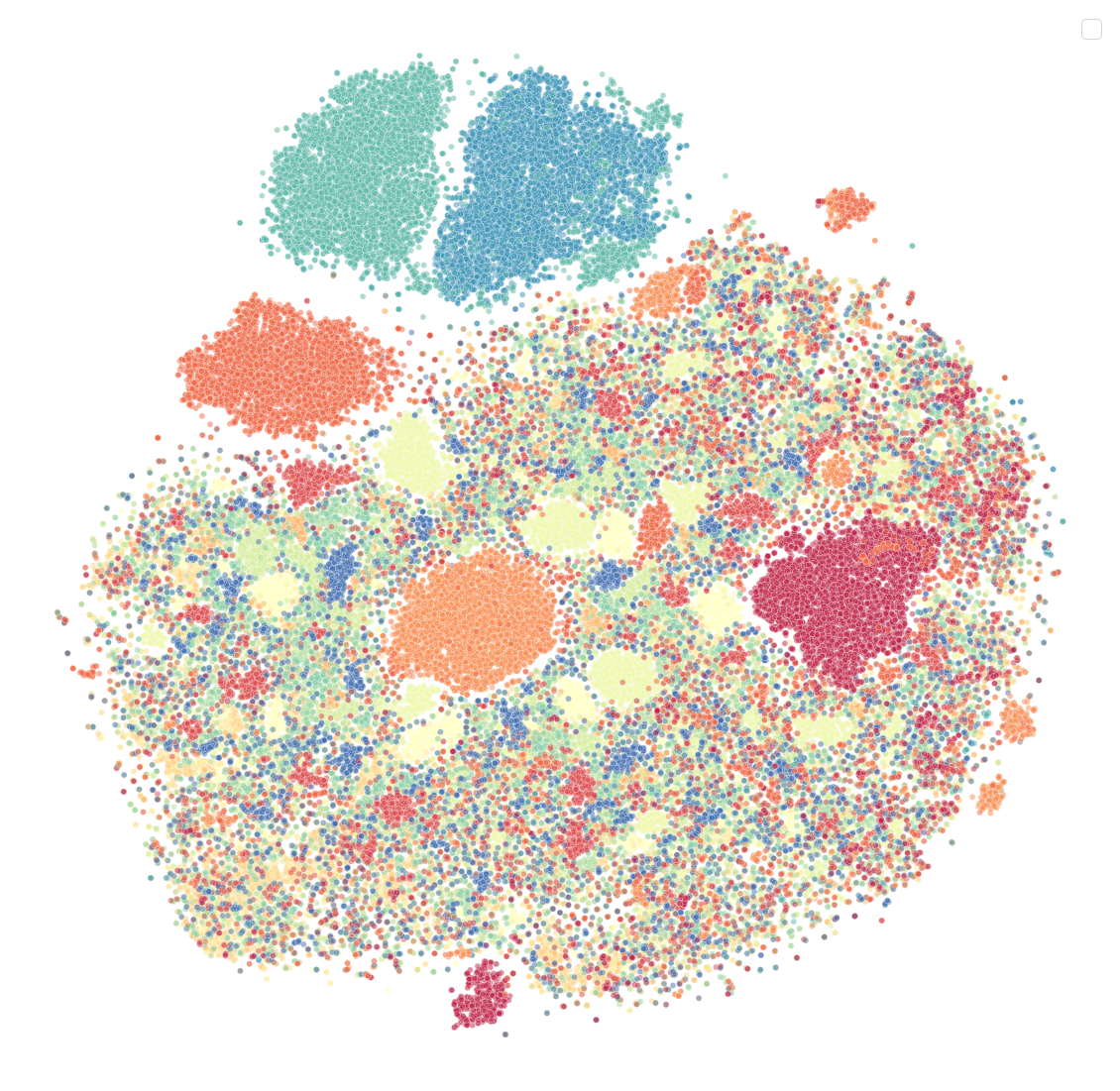}
    \caption{Cross-Lingual Transfer}
    \end{subfigure}
    \hfill
    \begin{subfigure}{0.49\textwidth}
    \includegraphics[width=\textwidth]{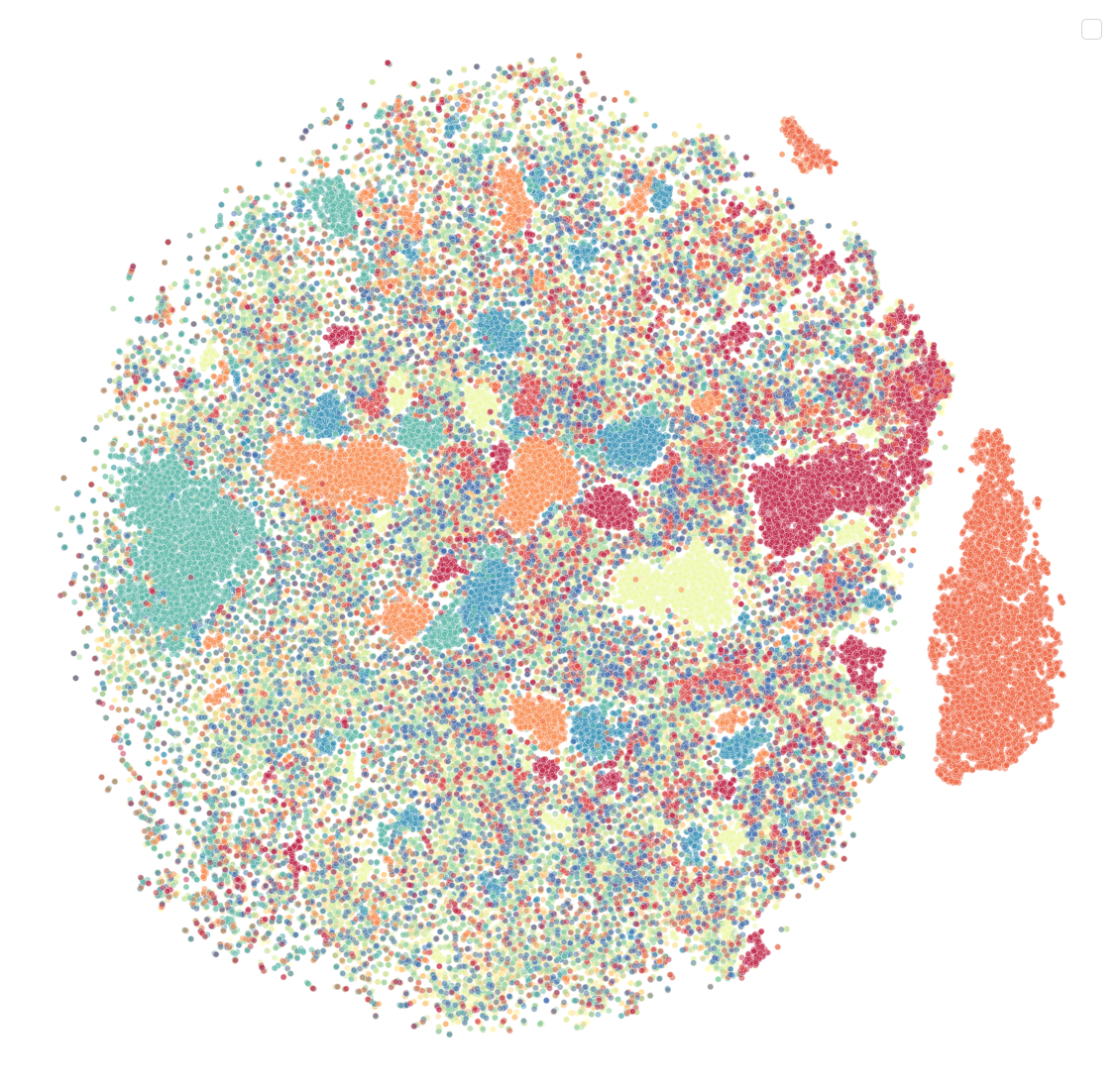}
    \caption{Translate-Train-$n$}
    \end{subfigure}
    \begin{subfigure}{0.49\textwidth}
    \includegraphics[width=\textwidth]{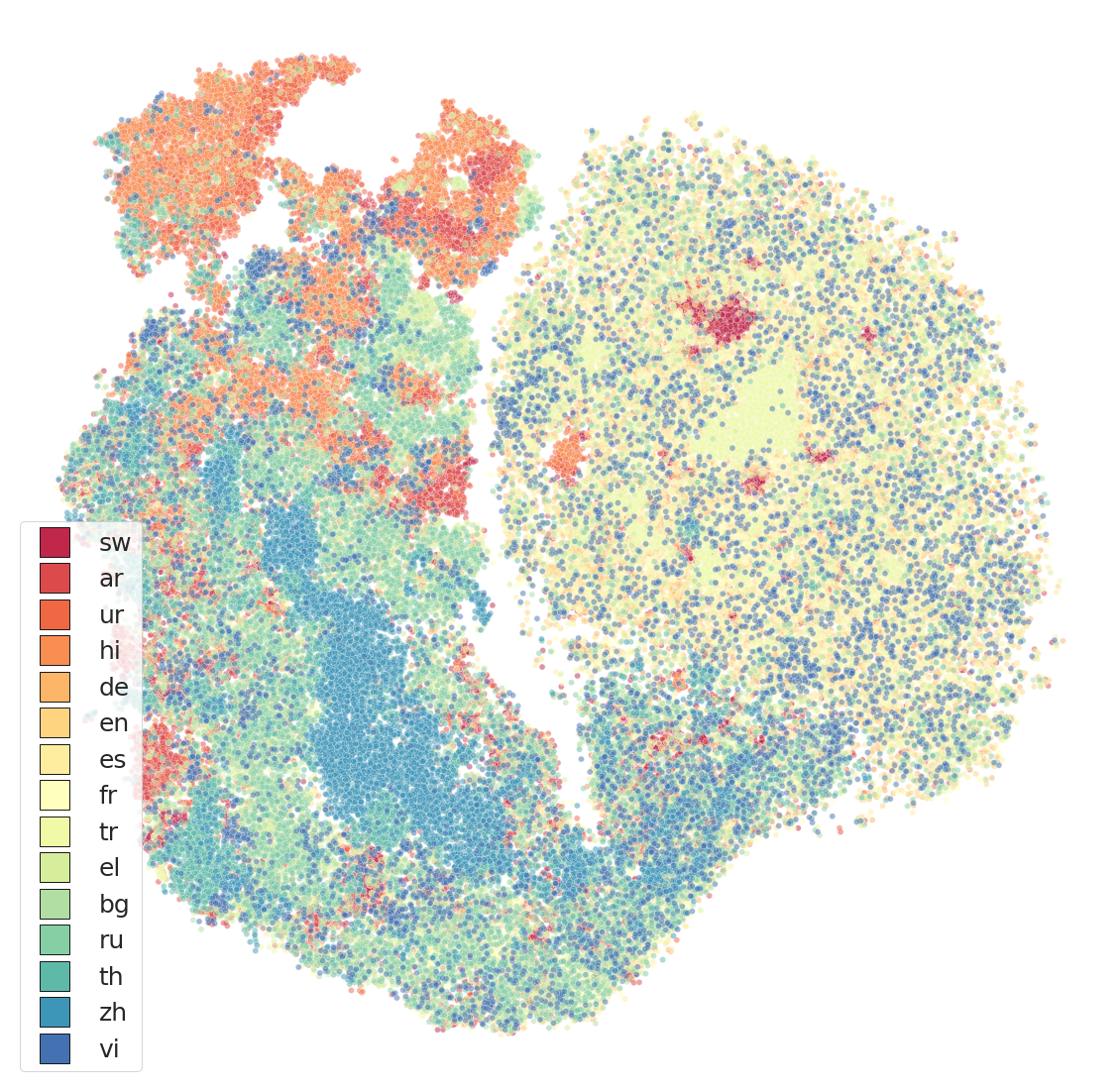}
    \caption{Domain Adversarial Neural Network}
    \end{subfigure}
    \hfill
    \begin{subfigure}{0.49\textwidth}
    \includegraphics[width=\textwidth]{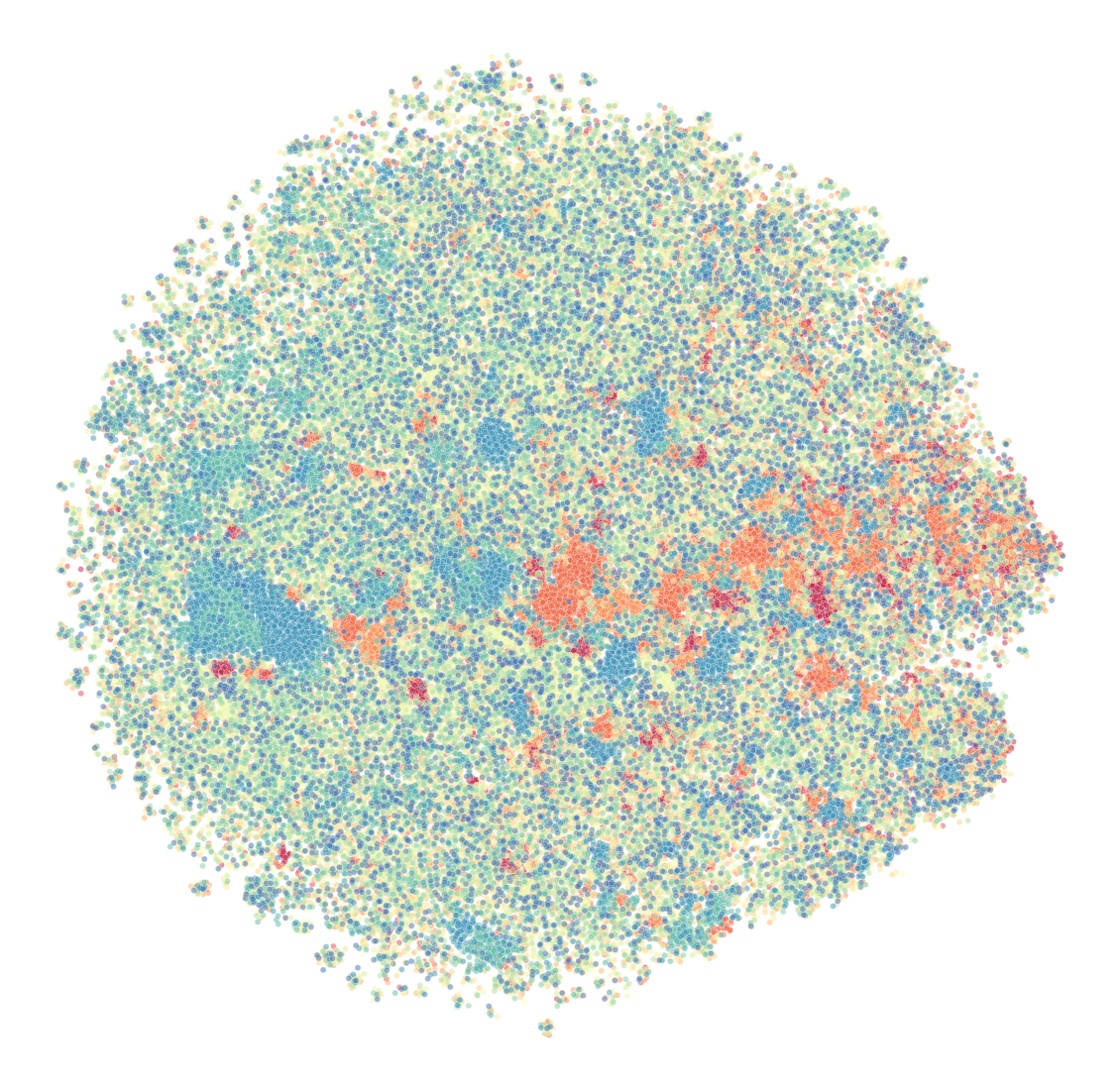}
    \caption{Code-Mixed Adversarial Training}
    \end{subfigure}
    \caption{t-SNE visualizations of XLM-R$_\text{base}$ representations fine-tuned using different methods. Here, we average all the token embeddings in the sentence instead of just using the \texttt{<s>} embedding.}
    \label{fig:tsne_avg_app}
\end{figure*}

%\begin{figure*}[h]
%\small
%    \centering
%    \includegraphics[width=0.75\textwidth]{figs/mlm_tsne_xnli_sb.png}
%    \caption{t-SNE visualization of the pretrained XLM-R$_\text{base}$ representation.}
%    \label{fig:tsne_pretrained}
%\end{figure*}

\end{document}